\documentclass{article} 
\usepackage{iclr2022_conference,times}

\usepackage{amsmath,amsfonts,bm}

\def\eqref#1{equation~\ref{#1}}

\def\1{\bm{1}}

\DeclareMathAlphabet{\mathsfit}{\encodingdefault}{\sfdefault}{m}{sl}
\SetMathAlphabet{\mathsfit}{bold}{\encodingdefault}{\sfdefault}{bx}{n}

\usepackage{hyperref}
\usepackage{url}
\usepackage{titlesec}

\usepackage[utf8]{inputenc} 
\usepackage[T1]{fontenc}    
\usepackage{hyperref}       
\usepackage{url}            
\usepackage{booktabs}       
\usepackage{amsfonts}       
\usepackage{nicefrac}       
\usepackage{microtype}      
\usepackage{graphicx}       
\usepackage[ruled,vlined]{algorithm2e}

\SetCommentSty{mycommfont}
\usepackage{float}
\usepackage{amsmath}
\usepackage{bbm}
\usepackage{cleveref}
\usepackage{color}
\usepackage{caption}
\usepackage{subcaption}
\usepackage{tabularx}
\usepackage{multirow}
\usepackage{array}
\newcolumntype{P}[1]{>{\centering\arraybackslash}p{#1}}
\newcolumntype{M}[1]{>{\centering\arraybackslash}m{#1}}
\usepackage{pifont}
\usepackage{xcolor}
\newcommand{\cmark}{\textcolor{green}{\ding{51}}}
\newcommand{\xmark}{\ding{55}}

\newcommand\algname{GradSign\xspace}

\newcommand{\ours}[1]{{\textcolor{brown}{#1}}}
\newtheorem{theorem}{Theorem}

\newtheorem{lemma}[theorem]{Lemma}

\titlespacing\section{0pt}{1pt plus 2pt minus 2pt}{0pt plus 2pt minus 2pt}
\titlespacing\subsection{0pt}{1pt plus 2pt minus 2pt}{0pt plus 2pt minus 2pt}
\titlespacing\subsubsection{0pt}{1pt plus 2pt minus 2pt}{0pt plus 2pt minus 2pt}
\title{\algname: Model Performance Inference with Theoretical Insights}

\author{  
 Zhihao Zhang \\
  Carnegie Mellon University\\
  \texttt{zhihaoz3@cs.cmu.edu}
  \And
  Zhihao Jia \\
  Carnegie Mellon University\\
  \texttt{zhihao@cmu.edu} \\
}

\iclrfinalcopy
\begin{document}

\maketitle
\begin{abstract}
A key challenge in neural architecture search (NAS) is quickly inferring the predictive performance of a broad spectrum of neural networks to discover statistically accurate and computationally efficient ones. We refer to this task as model performance inference (MPI).
The current practice for efficient MPI is gradient-based methods that leverage the gradients of a network at initialization to infer its performance.
However, existing gradient-based methods rely only on heuristic metrics and lack the necessary theoretical foundations to consolidate their designs.
We propose \algname, an accurate, simple, and flexible metric for model performance inference with theoretical insights.
A key idea behind \algname is a quantity $\Psi$ to analyze the {\em sample-wise optimization landscape} of different networks.
Theoretically, we show that $\Psi$ is an upper bound for both the training and true population losses of a neural network under reasonable assumptions.
However, it is computationally prohibitive to directly calculate $\Psi$ for modern neural networks. 
To address this challenge, we design \algname, an accurate and simple approximation of $\Psi$ using the gradients of a network evaluated at a random initialization state.
Evaluation on seven NAS benchmarks across three training datasets shows that \algname generalizes well to real-world neural networks and consistently outperforms state-of-the-art gradient-based methods for MPI evaluated by Spearman's $\rho$ and Kendall's Tau.
Additionally, we have integrated \algname into four existing NAS algorithms and show that the \algname-assisted NAS algorithms outperform their vanilla counterparts by improving the accuracies of best-discovered networks by up to 0.3\%, 1.1\%, and 1.0\% on three real-world tasks. Code is available at \url{https://github.com/cmu-catalyst/GradSign}
\end{abstract}
\section{Introduction}
\label{sec:introduction}
As deep learning methods evolve, neural architectures have gotten progressively larger and more sophisticated~\citep{he2015deep, ioffe2015batch, Krizhevsky_2017,devlin2019bert, rumelhart1986learning, JMLR:v15:srivastava14a, kingma2017adam}, making it increasingly challenging to {\em manually} design model architectures that can achieve state-of-the-art predictive performance.
To alleviate this challenge, recent work has proposed several approaches to {\em automatically} discovering statistically accurate and computationally efficient neural architectures.
The most common approach is {\em neural architecture search} (NAS), which explores a comprehensive search space of potential network architectures that use a set of predefined network modules as basic building blocks. Recent work shows that NAS is able to discover architectures that outperform human-designed counterparts~\citep{liu2018progressive, zoph2016neural, pham2018efficient}.

A key challenge in NAS is quickly assessing the predictive performance of a diverse set of candidate architectures to discover performant ones. We refer to this task as {\em model performance inference} (MPI).
A straightforward approach to MPI is directly training each candidate architecture on a dataset until convergence and recording the achieved training loss and validation accuracy~\citep{frankle2018lottery, chen2020lottery, liu2018progressive, zoph2016neural}.
Though accurate, this approach is computationally prohibitive and cannot scale to large networks or datasets.

The current practice to efficient MPI is {\em gradient-based methods} that leverage the gradient information of a network at initialization to infer its predictive performance~\citep{lee2018snip, wang2020picking, tanaka2020pruning}.
Compared to directly measuring the accuracy of candidate networks on a training dataset, gradient-based methods are computationally more efficient since they only require evaluating a mini-batch of gradients at initialization.
However, existing gradient-based methods rely only on heuristic metrics and lack the necessary theoretical insights to consolidate their designs.

In this paper, we propose \algname, a simple yet accurate metric for MPI with theoretical foundations. \algname is inspired by analyzing the {\em sample-wise optimization landscape} of a network. \algname takes as inputs a mini-batch of sample-wise gradients evaluated at a random initialization point and outputs a statistical evidence of a network that highly correlates to its well-trained predictive performance measured by accuracy on the entire dataset.

Prior theoretical results \citep{allen2019convergence} show that the optimization landscape of a randomly initialized network is nearly convex and semi-smooth for a sufficiently large neighborhood.
To realize its potential for MPI, we generalize these results to sample-wise optimization landscapes and propose a quantity $\Psi$ to measure the density of sample-wise local optima in the convex areas around a random initialization point.
Additionally, we prove that both the training loss and generalization error of a network are proportionally upper bounded by $\Psi^2$ under reasonable assumptions.

Based on our theoretical results, we design \algname, an accurate and simple approximation of $\Psi$. Empirically, we show that \algname can also generalize to realistic setups that may violate our assumptions. In addition, \algname is efficient to compute and easy to implement as it uses only the sample-wise gradient information of a network at a random initialization point. 

Extensive evaluation of \algname on seven NAS benchmarks (i.e., NAS-Bench-101, NAS-Bench-201, and five design spaces of NDS) across three datasets (i.e., CIFAR-10, CIFAR-100, and ImageNet16-120) shows that \algname consistently outperforms existing gradient-based methods in all circumstances. 
Furthermore, we have integrated \algname into existing NAS algorithms and show that the \algname-assisted variants of these NAS algorithms lead to more accurate architectures.

\paragraph{Contributions.} This paper makes the following contributions:
\begin{itemize} 
    \item We provide a new perspective to view the overall optimization landscape of a network as a combination of sample-wise optimization landscapes. Based on this insight, we introduce a new quantity $\Psi$ that provides an upper bound on both the training loss and generalization error of a network under reasonable assumptions.
    
    \item To infer $\Psi$, we propose \algname, an accurate and simple estimation of $\Psi$. \algname enables fast and efficient MPI using only the sample-wise gradients of a network at initialization.
    
    \item  We empirically show that \algname generalizes to modern network architectures and consistently outperforms existing gradient-based MPI methods. Additionally, \algname can be directly integrated into a variety of NAS algorithms to discover more accurate architectures.
\end{itemize}
\section{Related Work}
\subsection{Model Performance Inference}
Table~\ref{tab:relatedwork} summarizes existing approaches to inferring the statistical performance of neural architectures.

\noindent{\bf Sample-based methods} assess the performance of a neural architecture by training it on a dataset.
Though accurate, sample-based methods require a surrogate training procedure to evaluate each architecture.
EconNAS~\citep{zhou2020econas} mitigates the cost of training candidate architectures by reducing the number of training epochs, input dataset sizes, resolution of input images, and model sizes.

\noindent{\bf Theory-based methods} leverage recent advances in deep learning theory, such as Neural Tangent Kernel~\citep{jacot2018neural} and Linear Region Analysis~\citep{serra2018bounding}, to assess the predictive performance of a network~\citep{chen2020lottery, mellor2021neural, park2020towards}.
In particular, NNGP~\citep{park2020towards} infers a network's performance by fitting its kernel regression parameters on a training dataset and evaluating its accuracy on a validation set, which alleviates the burden of training. 
As another example, \citet{chen2020lottery} utilizes the kernel condition number proposed in \citet{xiao2020disentangling}, which can be theoretically proved to correlate to training convergence rate and generalization performance. However, this theoretical evidence is only guaranteed for extremely wide networks with a specialized initialization mode. While the linear region analysis used in~\citet{mellor2021neural}, \citet{lin2021zen} and~\citet{ chen2020lottery} is easy to implement, such technique is only applicable to networks with ReLU activations~\citep{agarap2018deep}.

\noindent{\bf Learning-based methods} train a separate network (e.g., graph neural networks) to predict a network's accuracy~\citep{liu2018progressive, luo2020semi, dai2019chamnet, wen2020neural, chen2020fitting, siems2020bench}.
Though these learned models can achieve high accuracies on a specific task, this approach requires constructing a training dataset with sampled architectures for each downstream task.
As a result, existing learning-based methods are generally task-specific and computationally prohibitive.


\noindent{\bf Gradient-based methods} infer the statistical performance of a network by leveraging its gradient information at initialization, which can be easily obtained using an automated differentiation tool of today's ML frameworks, such as PyTorch \citep{paszke2017automatic} and TensorFlow \citep{abadi2016tensorflow}.
The weight-wise salience score computed by several pruning at initialization~\citep{lee2018snip, wang2020picking, tanaka2020pruning} methods can easily be adapted to MPI settings by summing scores up. Though lack of theoretical foundations, such migrations have been empirically proven to be effective as baselines in recent works \citep{abdelfattah2021zero, mellor2021neural, lin2021zen}. An alternative stream of work~\citep{turner2019blockswap, turner2021neural, theis2018faster} uses approximated second-order gradients, known as empirical Fisher Information Matrix (FIM), at a random initialization point to infer the performance of a network.
Empirical FIM~\citep{martens2014new} is a valid approximation of a model's predictive performance only if the model's parameters are a Maximum Likelihood Estimation (MLE).
However, this assumption is invalid at a random initialization point, making FIM-based algorithms inapplicable.
A key difference between \algname and existing gradient-based methods is that \algname is based on a fine-grained analysis of sample-wise optimization landscapes rather than heuristic insights.
In addition, \algname also provides the first attempt for MPI by leveraging the optimization landscape properties contained in sample-wise gradient information, while prior gradient-based methods only focus on gradients evaluated in a full batch fashion.

\begin{table}
\caption{\label{tab:relatedwork}A summary of existing methods for model performance inference. 
The right four columns show (1) whether a method is based on theoretical results, (2) whether a method avoids expensive training process, (3) whether a method is applicable to different model architectures, and (4) whether a method is applicable across different tasks.
}
\begin{center}
\small
\begin{tabular}{|l|l|c|c|c|c|}
 \hline
  & \multirow{2}{*}{\bf Methods} & {\bf Theoretical} & {\bf Training}  & {\bf Model} & {\bf Task} \\
  & & {\bf Insight} & {\bf Free} & {\bf Independent} & {\bf Independent} \\
 \hline
{\bf Sample-Based} & EconNAS & \xmark & \xmark & \cmark & \cmark \\ 
\hline
\multirow{2}{*}{\bf Theory-Based}   & NNGP, TE-NAS, & \cmark & \cmark & \xmark & \cmark \\
& NASWOT, ZenNAS & & & &\\
 \hline
\multirow{2}{*}{\bf Learning-Based} & Neural Predictor,  & \xmark & \xmark & \cmark & \xmark \\
& One-Shot-NAS-GCN & & & &\\
 \hline
\multirow{3}{*}{\bf Gradient-Based} & Snip, Grasp, Synflow & \xmark & \cmark & \cmark & \cmark\\
& Fisher & & & &\\
 \cline{2-6}
  & \ours{GradSign (this paper)} & \cmark & \cmark & \cmark & \cmark \\
 \hline
\end{tabular}
\end{center}
\end{table}

\subsection{Neural Architecture Search}
Recent work \citep{he2021automl, cai2019once, cai2018proxylessnas, pmlr-v97-tan19a, Howard_2019_ICCV} has proposed several algorithms to explore a NAS search space and discover highly accurate networks.
RS~\citep{bergstra2012random} is one of the baseline algorithms that generates and evaluates architectures randomly in the search space. 
REINFORCE~\citep{williams1992simple} moves a step forward by reframing NAS as a reinforcement learning task where accuracy is the reward and architecture generation is the policy action. Given limited computational resources, BOHB~\citep{falkner2018bohb} uses Bayesian Optimization (BO) to propose candidates while uses HyperBand(HB)~\citep{li2017hyperband} for searching resource allocation. 
REA~\citep{real2019regularized} uses a simple yet effective evolutionary searching strategy that achieves state-of-the-art performance.
\algname is complementary to and can be combined with existing NAS algorithms.
We integrate \algname into the NAS algorithms mentioned above and show that \algname can consistently assist these NAS algorithms to discover more accurate architectures on various real-world tasks.

\subsection{Optimization Landscape Analysis}
Inspired by the fact that over-parameterized networks always find a remarkable fit for a training dataset~\citep{zhang2016understanding},
optimization landscape analysis has been one of the main focuses in deep learning theory~\citep{brutzkus2017globally, du2018gradient, ge2017learning, li2017convergence, soltanolkotabi2017learning, allen2019convergence}.
Even though existing theoretical results for optimization landscape analysis rely on strict assumptions on the landscape's smoothness, convexity, and initialization point, we can leverage theoretical insights to guide the design of \algname.
In addition, SGD-based optimizers trained from randomly initialized points hardly encounter non-smoothness or non-convexity in practice for a variety of architectures~\citep{goodfellow2014qualitatively}. Furthermore, \citet{allen2019convergence} provides theoretical evidence that for a sufficiently large neighborhood of a randomly initialized point, the optimization landscape is nearly convex and semi-smooth.
Different from existing optimization landscape analyses depending on 
objectives evaluated across a mini-batch of training samples, we propose a new perspective that decomposes a mini-batch objective into the aggregation of sample-wise optimization landscapes.
To the best of our knowledge, our work is the first attempt to MPI by leveraging sample-wise optimization landscapes.

\section{Theoretical Foundations}
\label{sec:theory}
\subsection{Insights}
\begin{figure}
     \centering
     \begin{subfigure}[b]{0.4\textwidth}
         \centering
         \includegraphics[width=\textwidth]{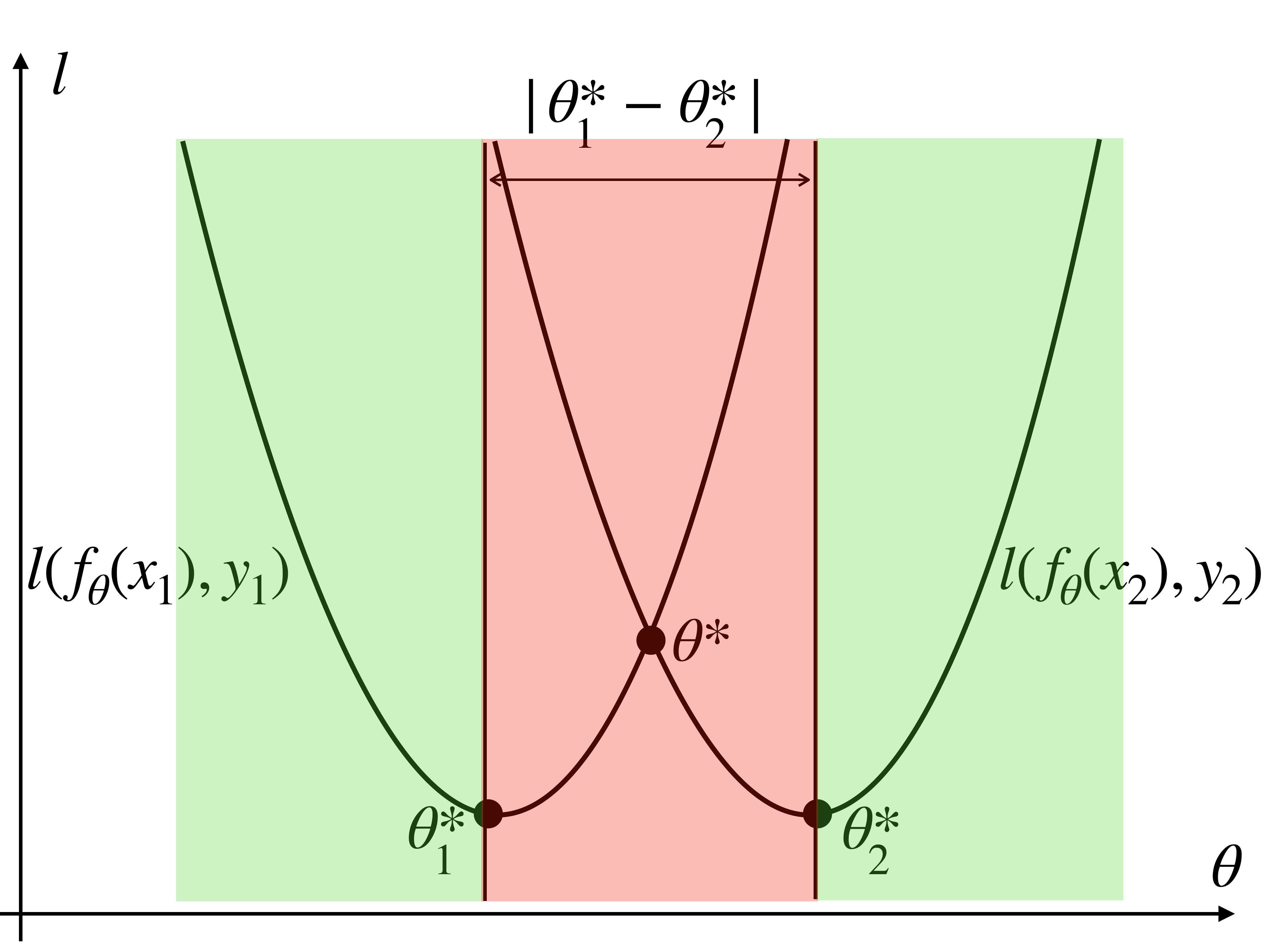}
         \caption{Optimization landscape with \textbf{sparser} sample-wise local optima corresponding to \textbf{worse} $J(\theta^*)$.}
         \label{fig:ill_1}
     \end{subfigure}
     \hfill
     \begin{subfigure}[b]{0.4\textwidth}
         \centering
         \includegraphics[width=\textwidth]{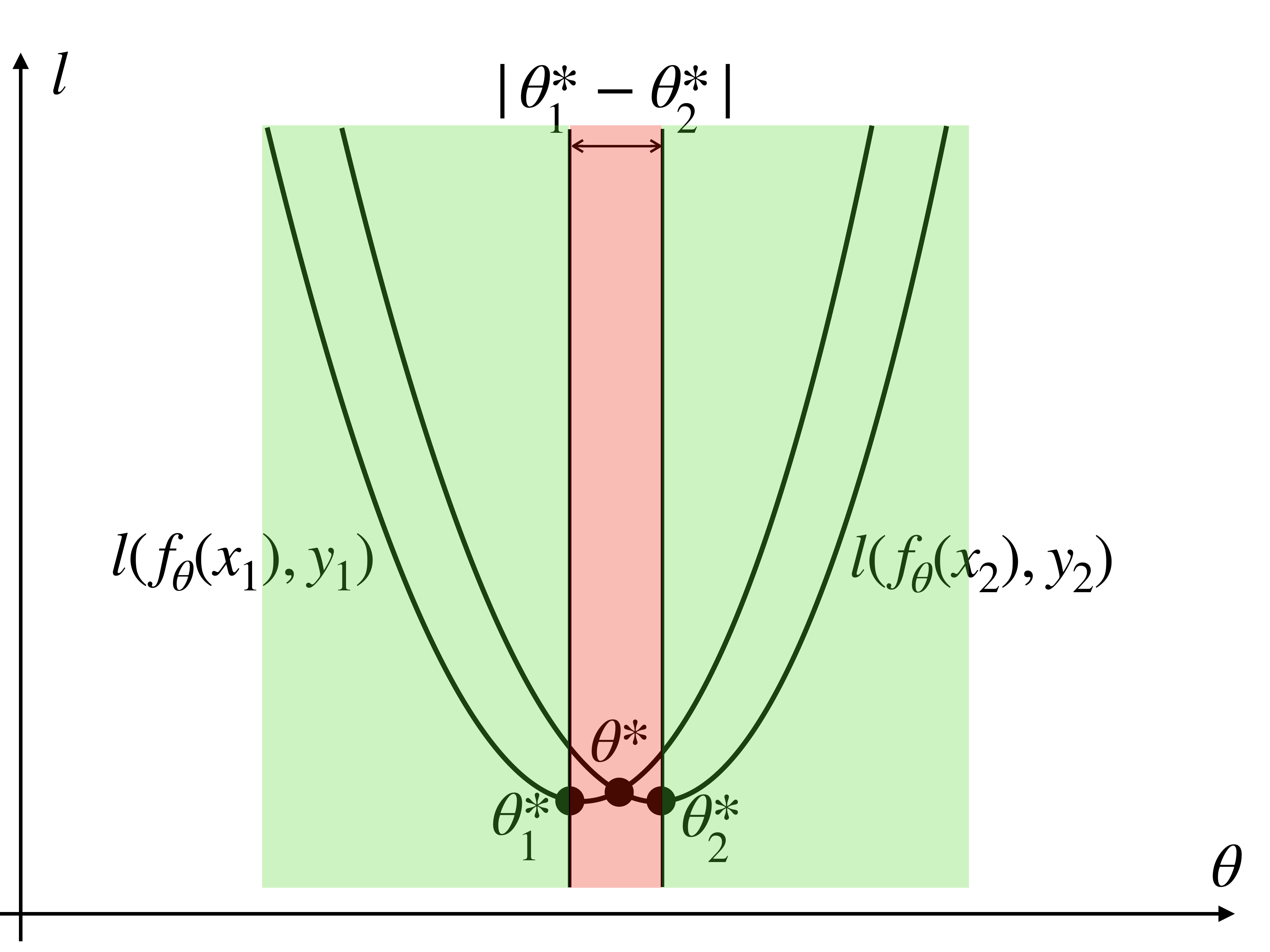}
        \caption{Optimization landscape with \textbf{denser} sample-wise local optima corresponding to \textbf{better} $J(\theta^*)$.}
         \label{fig:ill_2}
     \end{subfigure}
    \caption{Illustration of our theoretical insight that denser sample-wise local optima indicate lower training losses. As the distances ($|\theta_{1}^{*}-\theta_{2}^{*}|$, shown in red) between the local optima across samples reduce, there is a higher probability that the gradients of different samples have the same sign at a random initialization point, shown as the green areas.}
    \label{fig:ill}
\end{figure}

Conventional optimization landscape analyses focus on objectives across a mini-batch of training samples and miss potential evidence hidden in the optimization landscapes of individual samples.
By decomposing a mini-batch objective into the summation of {\em sample-wise} objectives across individual samples in a mini-batch, we can distinguish better local optima as illustrated in~\Cref{fig:ill}.
Both \Cref{fig:ill_1} and~\Cref{fig:ill_2} reach a local optimum at $\theta^*$ for the mini-batch objective $J=\frac{1}{2}(l(f_{\theta^*}(x_1), y_1)+l(f_{\theta^*}(x_2), y_2))$. However, the optimization landscape in \Cref{fig:ill_2} contains a better local optimum $\theta^*$ (i.e., a lower $J$).
This can be distinguished by analyzing the relative distance between local optima across training samples (i.e., $|\theta_1^*-\theta_2^*|$ in~\Cref{fig:ill}). 

For a mini-batch with more than two samples, we use a sample-wise local optima density measurement $\Psi$ defined in \Cref{sec:notation} to represent the overall closeness of sample-wise local optima. 
Intuitively, as the distances between the local optima across samples reduce (shown as the red areas in \Cref{fig:ill}), there is a higher probability that the gradients of different samples evaluated at a random initialization point have the same sign (shown as the green areas in \Cref{fig:ill}).
Driven by this insight, we propose \algname to infer the sample-wise local optima density $\Psi$ statistically.
The design of \algname is based on our theoretical results that a network with denser sample-wise local optima has lower training and generalization losses under reasonable assumptions.
We introduce the notations and assumptions in \Cref{sec:notation}, provide a formal derivation of our theoretical results in \Cref{sec:theorem}, and present \algname in \Cref{sec:metric}.

\subsection{Preliminaries}
\label{sec:notation}
We use $\mathcal{S}=\{(x_i, y_i)\}_{i\in[n]}$ to denote training samples, where each $x_i \in \mathbb{R}^d$ is a feature vector, and $y_i$ is the corresponding label. We use $l(\hat{y_i}, y_i)$ to represent a loss function where $\hat{y_i}$ is the prediction of our model. 
We use $f_\theta(\cdot):\mathbb{R}^d\rightarrow \mathbb{R}^o$ to denote the model parameterized by $\theta$ and use $\theta_0 \in \mathbb{R}^m$ to denote random initialized parameters where $m$ is the number of model parameters. Bold font constant denotes a constant vector such as $\mathbf{0}=[0,0,\cdots, 0]$ whose dimension depends on the corresponding situation. $\mathcal{D}$ denotes the underline data distribution, which is the same for training and testing.

Our theoretical results rely on an assumption that there exists a neighborhood $\Gamma_{\theta_0}$ for a random initialization point $\theta_0$ in which sample-wise optimization landscapes are almost convex and semi-smooth.
Note that our assumption is weaker than that of~\citep{allen2019convergence} since their analysis is focusing on the overall optimization landscape while we only consider the sample-wise optimization landscape which is a simpler case.
We use $\{\theta_i^*\}_{i\in[n]}\in \Gamma_{\theta_0} $ to denote a local optima in the convex areas attached to the $i$-th sample near the initialization point $\theta_0$. 
Under this assumption, the overall optimization landscape is also convex and semi-smooth within the neighborhood $\Gamma_{\theta_0}$ as additive operations preserve both. We use $\theta^*$ to denote a local optimum within $\Gamma_{\theta_0}$ for the mini-batch optimization landscape. 
Note that $\theta^*$ always lie in the convex hull of $\{\theta_i^*\}_{i\in[n]}$. 
Second, we assume that only gradient-based optimizers~\footnote{Gradient descent with infinitesimal step size} are used during training. Thus the optimizer eventually converges to $\theta^*$. 
Third, our theoretical analysis assumes that every Hessian in the set $\{\nabla^2 l(f_\theta(x_i), y_i)|\forall i \in [n], \theta \in \Gamma_{\theta_0}\}$ is almost diagonal as in the Neural Tangent Kernel (NTK) regime~\citep{jacot2018neural}. \Cref{sec:exp} shows that our method generalizes well to real-world networks that may violate this assumption.

\noindent\textbf{Sample-wise local optima density.}
We use {\em sample-wise local optima density} to represent the relative closeness of $\{\theta_i^*\}_{i\in[n]}$.
Given a dataset $\mathcal{S}=\{(x_i, y_i)\}_{i\in[n]}$, an objective function $l(\hat{y_i}, y_i)$, and a model class $f_\theta(\cdot)$, we use $\Psi_{\mathcal{S}, l}(f_{\theta_0}(\cdot))$ to measure the average distance 
between the local optima across samples $\{\theta_i^*\}_{i\in[n]}$ near a random initialization point $\theta_0$:
\begin{eqnarray}
\Psi_{\mathcal{S}, l}(f_{\theta_0}(\cdot)) &=& \frac{\sqrt{\mathcal{H}}}{n^2} \sum_{i, j} \|\theta_i^*-\theta_j^*\|_1
\end{eqnarray}
$\mathcal{H}\in \mathbb{R}$ is a smoothness upper bound: $\forall k \in [m], i \in [n], [\nabla^2 l(f_\theta(x_i), y_i)]_{k,k} \leq \mathcal{H}$. This upper bound always exists due to the smoothness assumption. Intuitively, $\Psi_{\mathcal{S}, l}(f_{\theta_0}(\cdot))$ can be interpreted as the mean Manhattan distance with respect to each pair of $\{\theta_i^*\}_{i\in[n]}$ normalized by the inverse of the square root of the smoothness upper bound.
The denser $\{\theta_i^*\}_{i\in[n]}$ are, the smaller  $\Psi_{\mathcal{S}, l}(f_{\theta_0}(\cdot))$ is.
In an ideal case, $\Psi_{\mathcal{S}, l}(f_{\theta_0}(\cdot)) = 0$ when all local optima are located at the same point.

%

\subsection{Main Results}
\label{sec:theorem}
We show the local optimum property of sample-wise optimization landscapes in the following lemma.

\begin{lemma}
\label{lemma1}
There exists no saddle point in a sample-wise optimization landscape and every local optimum is a global optimum.
\end{lemma}

Using Lemma~\ref{lemma1}, we can draw a relation between the training error $J=\frac{1}{n}\sum_{i}l(f_\theta(x_i), y_i)$ and $\Psi_{\mathcal{S}, l}(f_{\theta_0}(\cdot))$ using the following theorem.

\begin{theorem}
The training error of a network on a dataset $J=\frac{1}{n}\sum_{i}l(f_\theta(x_i), y_i)$ is upper bounded by $\frac{n^3}{2} \Psi_{\mathcal{S}, l}^2(f_{\theta_0}(\cdot))$, and the bound is tight when $\Psi_{\mathcal{S}, l}(f_{\theta_0}(\cdot))=0$.
\end{theorem}

Finally, we show that $\Psi_{\mathcal{S}, l}(f_{\theta_0}(\cdot))$ also provides an upper bound for the generalization performance of a network measured by population loss.

\begin{theorem}
Given that $\text{Var}_{(x_u, y_u)\sim \mathcal{D}}[\|\theta^*-\theta_u^*\|_1^2]$ is bounded by $\sigma^2$ where $\theta_u^*$ is a local optimum attached to the convex area near $\theta_0$ for $l(f_\theta(x_u), y_u)$. With probability $1-\delta$, the true population loss is upper bounded by $\frac{n^3}{2} \Psi_{\mathcal{S}, l}^2(f_{\theta_0}(\cdot)) + \frac{\sigma}{\sqrt{n\delta}}$.
\end{theorem}

A formal proof of all theoretical results is available in \Cref{proof:lemma_1}.

\noindent{\bf Main takeaways:}
A key takeaway of our theoretical results is that $\Psi_{\mathcal{S}, l}(f_{\theta_0}(\cdot))$ closely relates to an upper bound of the training and generalization performance of a network.
Albeit theoretically sound, $\Psi_{\mathcal{S} l}(f_{\theta_0}(\cdot))$ is intractable to be directly measured. Instead, we derive a simple yet accurate metric to reflect $\Psi_{\mathcal{S}, l}(f_{\theta_0}(\cdot))$, which we present in the next section.

\section{\algname}
\label{sec:metric}
Inspired by the theoretical results derived above, we introduce \algname, a simple yet accurate metric for model performance inference.
The key idea behind \algname is a quantity to statistically reflect the relative value of $\Psi_{\mathcal{S}, l}(f_{\theta_0}(\cdot))$. Specifically,
\begin{equation}\label{eq:core}
    \Psi_{\mathcal{S}, l}(f_{\theta_0}(\cdot)) \propto C-\sum_{k}\sum_{i,j}P(\text{sign}([\nabla_\theta l(f_\theta(x_i), y_i)|_{\theta_0}]_{k})=\text{sign}([\nabla_\theta l(f_\theta(x_j), y_j)|_{\theta_0}]_{k}))
\end{equation}
where $C$ is a constant and $k$ is the vector index. Detailed derivation is given in \Cref{eq:final}. Using the above relation, we can infer $\Psi_{\mathcal{S}, l}(f_{\theta_0}(\cdot))$ 
by directly measuring the signs of the gradients for a mini-batch of training samples at a randomly initialized point instead of going through an end-to-end training process.


To enable more efficient calculation of $\Psi_{\mathcal{S}, l}(f_{\theta_0}(\cdot))$, we make a further simplification and  use the following sample observation:
\begin{equation}
    \sum_{k}|\sum_{i}\text{sign}([\nabla_\theta l(f_\theta(x_i), y_i)|_{\theta_0}]_{k})|
\end{equation}
to infer the true probability $\sum_{k}\sum_{i,j}P(\text{sign}([\nabla_\theta l(f_\theta(x_i), y_i)|_{\theta_0}]_{k})=\text{sign}([\nabla_\theta l(f_\theta(x_j), y_j)|_{\theta_0}]_{k}))$ whose relation is given by:
\begin{eqnarray}
&&\frac{1}{n^2}\sum_{i,j}\mathbbm{1}_{\text{sign}([\nabla_\theta l(f_\theta(x_i), y_i)|_{\theta_0}]_{k})=\text{sign}([\nabla_\theta l(f_\theta(x_j), y_j)|_{\theta_0}]_{k})}
\\
&\propto& |\sum_{i}\text{sign}([\nabla_\theta l(f_\theta(x_i), y_i)|_{\theta_0}]_{k})|^2
\end{eqnarray}
 The proof of this simplification is included in \Cref{eq:sim}. Given the above relationship, we formally state our algorithm pipeline in \Cref{alg:GradSign}. Note that a higher \algname score indicates better model performance as we have an inverse correlation in \Cref{eq:core}.
 
 \begin{algorithm}[H]
\SetAlgoLined
\KwResult{\algname score $\tau_f$ for a function class $f_\theta$}
 Given $\mathcal{S}=\{(x_i, y_i)\}_{i \in [n]}$, randomly select initialization point $\theta_0$\;
 Initialize $g[n, m]$\;
 \For{$i=1,2,\cdots, n$}{
  \For{$k=1,2,\cdots, m$}{
    $g[i, k] = \text{sign}([\nabla_\theta l(f_\theta(x_i), y_i)|_{\theta_0}]_k)$
  }
 }
 $\tau_f =  \sum_{k}|\sum_{i}g[i, k]|$\;
 \textbf{return} $\tau_f$
 \caption{\algname}
 \label{alg:GradSign}
\end{algorithm}

\section{Experiments}
\label{sec:exp}

In this section, we empirically verify the effectiveness of our metric against existing gradient-based methods on three neural architecture search (NAS) benchmarks, including NAS-Bench-101~\citep{ying2019bench}, NAS-Bench-201~\citep{dong2020bench} and NDS~\citep{radosavovic2019network}\footnote{All datasets have consented for research purposes and no identifiable personal information is included.}.
Theory-based, sample-based, and learning-based methods are excluded in our evaluation, as they either require further training processes or have strong assumptions not suitable for generic architectures.

\noindent{\bf Baselines.} We compare \algname against existing gradient-based methods, including snip~\citep{lee2018snip}, grasp~\citep{wang2020picking}, fisher~\citep{turner2019blockswap}, and Synflow.
In addition, we also include grad\_norm as a heuristic method and a one-shot MPI metric NASWOT\citep{mellor2021neural}.
Since all gradient-based methods share a similar calculation pipeline (i.e., evaluating the gradients of a mini-batch at a random initialization point), we set the initialization mode and batch size to be the same across all methods to guarantee fairness. Experimental setup details are included in \Cref{app_exp}. 
To align with the experimental setup of prior work~\citep{abdelfattah2021zerocost,mellor2021neural},
we use two criteria to evaluate the correlations between different metrics and test accuracies across approximately 20k networks:

\noindent\textbf{Spearman's $\rho$}~\citep{daniel1990applied} characterizes the monotonic relationships between two variables. The correlation score is restricted in range [-1, 1], where $\rho=1$ denotes a perfect positive monotonic relationship and $\rho=-1$ denotes a perfect negative monotonic relationship. 
Following prior work, we use Spearman's $\rho$ to evaluate gradient-based methods on NAS-Bench-101 and NAS-Bench-201. 

\noindent\textbf{Kendall's Tau}. Similar to Spearman's $\rho$ as a correlation measurement, Kendall's Tau is also restricted between [-1, 1]. While Spearman's $\rho$ is more sensitive to error and discrepancies, Kendall's Tau is more robust with a smaller gross error sensitivity. We use Kendall's Tau to quantify the correlation between one-shot metric scores and model testing accuracies over the NDS search space.
\subsection{NAS-Bench-101}
\label{sec:101}
NAS-Bench-101 is the first dataset targeting large-scale neural architecture space,  containing 423k unique convolutional architectures trained on the CIFAR-10 dataset. The benchmark provides the test accuracy of each architecture in the search space, which we use to calculate the corresponding Spearman's $\rho$. 
We use a randomly sampled subset with approximately $4500$ architectures of the original search space and a batch size of 64 in this experiment.
\Cref{tab:NAS-101} summarizes the results. 
GradSign significantly outperforms existing gradient-based methods and heuristic approaches and improves the Spearman's $\rho$ score by $25\%$ compared to the best existing method ($0.363\rightarrow 0.449$).
\setlength{\tabcolsep}{3pt}
\begin{table}[H]
\caption{\label{tab:NAS-101}Performance of existed MPI methods (gradient-based + NASWOT) on NAS-Bench-101 evaluated by Spearman's $\rho$.}
\begin{center}
\begin{tabular}{l|ccccccc}
\hline
Dataset        & \textit{grad\_norm} & snip  & grasp & fisher & Synflow & NASWOT & \ours{GradSign} \\ \hline
CIFAR10        & 0.263     & 0.189 & 0.315  & 0.3   & 0.363   & 0.324      & \textbf{0.449}    \\ \hline
\end{tabular}
\end{center}
\end{table}
\subsection{NAS-Bench-201}
\label{sec:201}
NAS-Bench-201 is an extended version of NAS-Bench-101 with a different search space, containing 15,625 cell-based candidate architectures evaluated across three datasets: CIFAR-10, CIFAR-100~\citep{Krizhevsky09learningmultiple} and ImageNet 16-120~\citep{russakovsky2015imagenet}. The benchmark provides the test accuracies on the three datasets for all candidate architectures in the search space.
We evaluate Spearman's $\rho$ scores for GradSign and existing gradient-based methods. The experiments were conducted overall 15,265 architectures in NAS-Bench-201. The batch size is set to 64. The results on the three datasets are summarized in \Cref{tab:NAS-201}. \algname consistently achieves the best performance across all three datasets and improves the Spearman's $\rho$ scores by $\approx 4\%$ over the best existing approaches.
This improvement is significant as the more Spearman's $\rho$ approaches $1$, the more difficult it can be further improved.
\setlength{\tabcolsep}{3pt}
\begin{table}[H]
\caption{\label{tab:NAS-201}Performance of existed MPI methods (gradient-based + NASWOT + ZenNAS) on NAS-Bench-201 evaluated by Spearman's $\rho$.}
\begin{center}
\begin{tabular}{lcccccccc}
\hline
Dataset   &    ZenNAS & \textit{grad\_norm} & snip  & grasp & fisher & Synflow & NASWOT & \ours{GradSign} \\ \hline
CIFAR10   &  -0.016   & 0.594      & 0.595 & 0.51  & 0.36   & 0.737   & 0.728      & \textbf{0.765}    \\
CIFAR100  &  -0.041   & 0.637      & 0.637 & 0.549 & 0.386  & 0.763   & 0.703      & \textbf{0.793}    \\
ImageNet16-120 &  0.032   & 0.579      & 0.579 & 0.552 & 0.328  & 0.751   & 0.696      & \textbf{0.783}    \\ \hline
\end{tabular}
\end{center}
\end{table}
We further select 1000 architecture candidates randomly in the NAS-Bench-201 search space and visualize their testing accuracies against \algname scores in \Cref{fig:201}. 
\Cref{fig:201} shows a highly positive correlation between the \algname score and actual test accuracy of 1000 architectures. A higher \algname score indicates higher confidence for the statistical performance of architecture.
Note that the \algname scores show a clustering pattern, which may correspond to different architecture classes in the NAS-Bench-201 search space.
\subsection{NAS Design Space (NDS)}
NDS is a unified searching framework that includes five different design spaces: NAS-Net~\citep{zoph2018learning}, AmoebaNet~\citep{real2019regularized}, PNAS~\citep{liu2018progressive}, ENAS~\citep{pham2018efficient}, DARTS~\citep{liu2018darts}. Each space contains approximately one thousand networks fully trained on the CIFAR-10 dataset. We include the performance of our method along with \textit{grad\_norm}, Synflow, and NASWOT on all five design spaces evaluated by Kendall's Tau and show the results in \Cref{tab:DARTS-Kendall}. \algname significantly and consistently outperforms all other MPI methods in all five design spaces. 
\begin{table}[H]
\caption{\label{tab:DARTS-Kendall} Performance of existed MPI methods on five design spaces in NDS trained over CIFAR-10 evaluated by Kendall's Tau.}
\begin{center}
\begin{tabular}{cccccc}
\hline
           & DARTS         & ENAS          & PNAS          & NASNet        & Amoeba        \\ \hline
\textit{grad\_norm} & 0.28          & -0.02         & -0.01         & -0.08         & -0.10         \\
Synflow    & 0.37          & 0.02          & 0.03          & -0.03         & -0.06         \\
NASWOT     & 0.48          & 0.34          & 0.31          & \textbf{0.31} & 0.20          \\
\ours{GradSign}  & \textbf{0.54} & \textbf{0.43} & \textbf{0.40} & \textbf{0.31} & \textbf{0.24} \\ \hline
\end{tabular}
\end{center}
\end{table}
\subsection{Architecture Selection}
We evaluate whether \algname can be directly used to select highly accurate architectures in a NAS search space.
To pick a top architecture, we randomly sample $N$ candidates in a NAS search space, choose the one with the highest \algname score, and measure its validation/test accuracies (mean$\pm$std).
We compare \algname with Synflow, NASWOT, {\tt Random}, and {\tt Optimal}, where {\tt Random} uniformly samples architectures in the search space, while {\tt Optimal} always chooses the best architecture across $N$ candidates.
The results\footnote{the results for NASWOT are referenced from their paper~\citep{mellor2021neural}} are summarized in Table~\ref{tab:NAS-201-TFree}. $N$ in parenthesis indicates the number of architectures sampled in each run. All methods can generally find more accurate architectures with a high $N$.
In addition to outperforming Synflow and NASWOT, \algname($N=100$) can also find better networks even compared to NASWOT($N=1000$).
The results show that \algname can directly identify accurate architectures besides highly correlating to networks' test accuracies.
\setlength{\tabcolsep}{1pt}
\begin{table}[H]
\caption{\label{tab:NAS-201-TFree} Mean $\pm$ std accuracy evaluated on NAS-Bench-201. All results are averaged over 500 runs.
All searches are conducted on CIFAR-10 while the selected architectures are evaluated on CIFAR-10, CIFAR-100, and ImageNet16-120. $N$ in parenthesis is the number of networks sampled in each run.}
\begin{center}
\begin{tabular}{ccccccc}
\hline
                       \multirow{2}{*}{Methods} & \multicolumn{2}{c}{CIFAR-10}                & \multicolumn{2}{c}{CIFAR-100}               & \multicolumn{2}{c}{ImageNet16-120}          \\ \cline{2-7} 
                                                & Validation           & Test                 & Validation           & Test                 & Validation           & Test                 \\ \hline
  Synflow(N=100)                 & \textbf{89.83$\pm$0.75} & 93.12$\pm$0.52          & 69.89$\pm$1.87          & 69.94$\pm$1.88          & 41.94$\pm$4.13          & 42.26$\pm$4.26          \\
 NASWOT(N=100)                   & 89.55$\pm$0.89          & 92.81$\pm$0.99          & 69.35$\pm$1.70          & 69.48$\pm$1.70          & 42.81$\pm$3.05          & 43.10$\pm$3.16          \\
  NASWOT(N=1000)                   & 89.69$\pm$0.73          & 92.96$\pm$0.81          & 69.98$\pm$1.22          & 69.86$\pm$1.21          & \textbf{44.44$\pm$2.10} & \textbf{43.95$\pm$2.05} \\
   \ours{GradSign(N=100)}              & \textbf{89.84$\pm$0.61} & \textbf{93.31$\pm$0.47} & \textbf{70.22$\pm$1.32} & \textbf{70.33$\pm$1.28} & 42.07$\pm$2.78          & 42.42$\pm$2.81          \\
   \hline
\tt{Random}               & 83.20$\pm$13.28          & 86.61$\pm$13.46          & 60.70$\pm$12.55         & 60.83$\pm$12.58          & 33.34$\pm$9.39 & 33.13$\pm$9.66 \\
 \tt{Optimal}(N=100)               & 91.05$\pm$0.28          & 93.84$\pm$0.23         & 71.45$\pm$0.79       & 71.56$\pm$0.78          & 45.37$\pm$0.61 & 45.67$\pm$0.64\\
 \hline
\end{tabular}
\end{center}
\end{table}
\subsection{\algname-Assisted Neural Architecture Search}
Besides evaluating \algname on the Spearman's $\rho$ and Kendall's Tau scores as prior work, we also integrate \algname into various neural architecture search algorithms and evaluate how \algname can assist neural architecture search on real-world tasks.
Specifically, we integrate \algname into four NAS algorithms: REA, REINFORCE, BOHB and RS.
We design a corresponding method for each NAS algorithm that uses the \algname scores of candidate architectures to guide the search.
Specifically, we integrate \algname into each NAS algorithm by replacing the random selection of architectures with \algname-assisted selection. We name these \algname-assisted variants G-REA, G-REINFROCE, G-HB, and G-RS, and describe their algorithm details in~\Cref{sec:appendix}.

To evaluate how \algname can improve the search procedure of NAS algorithms, we run each algorithm with and without \algname's assistance for 500 runs on NAS-Bench-201 and report the validation and test accuracies of the best-discovered architecture in each run.
Following prior work~\citep{mellor2021neural, dong2020bench}, all searches are conducted on the CIFAR-10 dataset with a time budget of 12000s while the performance is evaluated on CIFAR-10, CIFAR-100 and ImageNet16-120.
The baselines also include A-REA~\citep{mellor2021neural}, a variant of REA that uses the NASWOT scores at the initial population selection phase.



\Cref{tab:NAS-201-Algo} shows the results. The \algname-assisted NAS algorithms outperform their counterparts by improving test accuracy by up to 0.3\%, 1.1\%, and 1.0\% on the three datasets. 

\setlength{\tabcolsep}{3pt}
\begin{table}[H]
\caption{\label{tab:NAS-201-Algo} Mean $\pm$ std accuracy evaluated over NAS-Bench-201. All results are averaged over 500 runs. To make a fair comparison across all the methods, the search is performed on CIFAR-10 dataset while the architectures' performance are evaluated over CIFAR-10, CIFAR-100 and ImageNet16-120. All the methods have a search time budget of 12000s. Note that the benchmark results might not match with the original paper as we have run all the experiments from start in a environment different from \cite{dong2020bench}.}
\begin{center}
\begin{tabular}{ccccccc}
\hline
      \multirow{2}{*}{Methods}      & \multicolumn{2}{c}{CIFAR-10}                & \multicolumn{2}{c}{CIFAR-100}               & \multicolumn{2}{c}{ImageNet16-120}          \\ \cline{2-7} 
 & Validation           & Test                 & Validation           & Test                 & Validation           & Test                 \\ \hline
REA         & 91.08$\pm$0.45          & 93.85$\pm$0.44          & 71.59$\pm$1.33          & 71.64$\pm$1.25          & 44.90$\pm$1.20          & 45.25$\pm$1.41          \\
A-REA       & 91.20$\pm$0.27 & - & 71.95$\pm$0.99 & - & \textbf{45.70$\pm$1.05} & -  \\ 
\ours{G-REA}       & \textbf{91.27$\pm$0.58} & \textbf{94.10$\pm$0.52} & \textbf{72.64$\pm$1.57} & \textbf{72.70$\pm$1.50} & \textbf{45.69$\pm$1.33} & \textbf{45.7$\pm$1.32}  \\ \hline
RS          & 90.93$\pm$0.37          & 93.72$\pm$0.38          & 70.96$\pm$1.12          & 71.07$\pm$1.07          & 44.47$\pm$1.08          & 44.61$\pm$1.22          \\
\ours{G-RS}        & \textbf{91.24$\pm$0.21} & \textbf{94.02$\pm$0.21} & \textbf{72.15$\pm$0.77} & \textbf{72.20$\pm$0.76} & \textbf{45.38$\pm$0.79} & \textbf{45.77$\pm$0.79} \\ \hline
REINFORCE   & 90.32$\pm$0.89          & 93.21$\pm$0.82          & \textbf{70.03$\pm$1.75} & 70.14$\pm$1.73          & 43.57$\pm$2.09          & 43.64$\pm$2.24          \\
\ours{G-REINFORCE} & \textbf{90.47$\pm$0.55} & \textbf{93.37$\pm$0.47} & \textbf{70.00$\pm$1.20}          & \textbf{70.20$\pm$1.29} & \textbf{44.33$\pm$1.25} & \textbf{44.05$\pm$1.48} \\ \hline
BOHB        & 90.84$\pm$0.49          & 93.64$\pm$0.49          & 70.82$\pm$1.29          & 70.92$\pm$1.26          & 44.36$\pm$1.37          & 44.50$\pm$1.50          \\
\ours{G-HB}        & \textbf{91.18$\pm$0.26} & \textbf{93.96$\pm$0.25} & \textbf{71.92$\pm$0.92} & \textbf{71.99$\pm$0.85} & \textbf{45.29$\pm$0.84} & \textbf{45.53$\pm$0.92} \\ \hline
\end{tabular}
\end{center}
\end{table}

\section{Conclusion}
\label{sec:conclusion}
In this paper, we propose a model performance inference metric \algname and provide theoretical foundations to support our metric. Instead of focusing on full batch optimization landscape analysis, we move a step further to sample-wise optimization landscape properties, which give us additional information to uncover the quality of the local optima encountered on the optimization trajectory. We propose $\Psi_{\mathcal{S}, l}(f_{\theta_0}(\cdot))$ to quantitatively characterize the potential of a model $f_\theta(\cdot)$ at a random initialization point $\theta_0$ based on our theory results. Finally, we design the \algname metric to statistically infer the value of $\Psi_{\mathcal{S}, l}(f_{\theta_0}(\cdot))$ to give out our final score for model performance inference. Empirically, we have demonstrated that our method consistently achieves the best correlation with true model performance among all other gradient-based metrics. In addition, we also verified the practical value of our method in assisting existed NAS algorithms to achieve better results. Given that our metric is generic and promising, we believe that our work not only assists in accelerating MPI-related applications but sheds some light on optimization landscape analysis as well. Meanwhile, the effectiveness of our method may further reduce the energy cost introduced by modern NAS algorithms. In addition, one of the future work of \algname can be adding normalization across different architecture classes to tackle the clustering problem in~\Cref{fig:201}.

\bibliography{iclr2022_conference}

\begin{thebibliography}{64}
\providecommand{\natexlab}[1]{#1}
\providecommand{\url}[1]{\texttt{#1}}
\expandafter\ifx\csname urlstyle\endcsname\relax
  \providecommand{\doi}[1]{doi: #1}\else
  \providecommand{\doi}{doi: \begingroup \urlstyle{rm}\Url}\fi

\bibitem[Abadi et~al.(2016)Abadi, Barham, Chen, Chen, Davis, Dean, Devin,
  Ghemawat, Irving, Isard, et~al.]{abadi2016tensorflow}
Mart{\'\i}n Abadi, Paul Barham, Jianmin Chen, Zhifeng Chen, Andy Davis, Jeffrey
  Dean, Matthieu Devin, Sanjay Ghemawat, Geoffrey Irving, Michael Isard, et~al.
\newblock Tensorflow: A system for large-scale machine learning.
\newblock In \emph{12th $\{$USENIX$\}$ symposium on operating systems design
  and implementation ($\{$OSDI$\}$ 16)}, pp.\  265--283, 2016.

\bibitem[Abdelfattah et~al.(2021{\natexlab{a}})Abdelfattah, Mehrotra, Dudziak,
  and Lane]{abdelfattah2021zero}
Mohamed~S Abdelfattah, Abhinav Mehrotra, {\L}ukasz Dudziak, and Nicholas~D
  Lane.
\newblock Zero-cost proxies for lightweight nas.
\newblock \emph{arXiv preprint arXiv:2101.08134}, 2021{\natexlab{a}}.

\bibitem[Abdelfattah et~al.(2021{\natexlab{b}})Abdelfattah, Mehrotra, Dudziak,
  and Lane]{abdelfattah2021zerocost}
Mohamed~S. Abdelfattah, Abhinav Mehrotra, {\L}ukasz Dudziak, and Nicholas~D.
  Lane.
\newblock {Zero-Cost Proxies for Lightweight NAS}.
\newblock In \emph{International Conference on Learning Representations
  (ICLR)}, 2021{\natexlab{b}}.

\bibitem[Agarap(2018)]{agarap2018deep}
Abien~Fred Agarap.
\newblock Deep learning using rectified linear units (relu).
\newblock \emph{arXiv preprint arXiv:1803.08375}, 2018.

\bibitem[Allen-Zhu et~al.(2019)Allen-Zhu, Li, and Song]{allen2019convergence}
Zeyuan Allen-Zhu, Yuanzhi Li, and Zhao Song.
\newblock A convergence theory for deep learning via over-parameterization.
\newblock In \emph{International Conference on Machine Learning}, pp.\
  242--252. PMLR, 2019.

\bibitem[Bergstra \& Bengio(2012)Bergstra and Bengio]{bergstra2012random}
James Bergstra and Yoshua Bengio.
\newblock Random search for hyper-parameter optimization.
\newblock \emph{Journal of machine learning research}, 13\penalty0 (2), 2012.

\bibitem[Brutzkus \& Globerson(2017)Brutzkus and
  Globerson]{brutzkus2017globally}
Alon Brutzkus and Amir Globerson.
\newblock Globally optimal gradient descent for a convnet with gaussian inputs.
\newblock In \emph{International conference on machine learning}, pp.\
  605--614. PMLR, 2017.

\bibitem[Cai et~al.(2018)Cai, Zhu, and Han]{cai2018proxylessnas}
Han Cai, Ligeng Zhu, and Song Han.
\newblock Proxylessnas: Direct neural architecture search on target task and
  hardware.
\newblock \emph{arXiv preprint arXiv:1812.00332}, 2018.

\bibitem[Cai et~al.(2019)Cai, Gan, Wang, Zhang, and Han]{cai2019once}
Han Cai, Chuang Gan, Tianzhe Wang, Zhekai Zhang, and Song Han.
\newblock Once-for-all: Train one network and specialize it for efficient
  deployment.
\newblock \emph{arXiv preprint arXiv:1908.09791}, 2019.

\bibitem[Chen et~al.(2020{\natexlab{a}})Chen, Frankle, Chang, Liu, Zhang, Wang,
  and Carbin]{chen2020lottery}
Tianlong Chen, Jonathan Frankle, Shiyu Chang, Sijia Liu, Yang Zhang, Zhangyang
  Wang, and Michael Carbin.
\newblock The lottery ticket hypothesis for pre-trained bert networks.
\newblock \emph{arXiv preprint arXiv:2007.12223}, 2020{\natexlab{a}}.

\bibitem[Chen et~al.(2020{\natexlab{b}})Chen, Xie, Wu, Wei, Xu, and
  Tian]{chen2020fitting}
Xin Chen, Lingxi Xie, Jun Wu, Longhui Wei, Yuhui Xu, and Qi~Tian.
\newblock Fitting the search space of weight-sharing nas with graph
  convolutional networks.
\newblock \emph{arXiv preprint arXiv:2004.08423}, 2020{\natexlab{b}}.

\bibitem[Dai et~al.(2019)Dai, Zhang, Wu, Yin, Sun, Wang, Dukhan, Hu, Wu, Jia,
  et~al.]{dai2019chamnet}
Xiaoliang Dai, Peizhao Zhang, Bichen Wu, Hongxu Yin, Fei Sun, Yanghan Wang,
  Marat Dukhan, Yunqing Hu, Yiming Wu, Yangqing Jia, et~al.
\newblock Chamnet: Towards efficient network design through platform-aware
  model adaptation.
\newblock In \emph{Proceedings of the IEEE/CVF Conference on Computer Vision
  and Pattern Recognition}, pp.\  11398--11407, 2019.

\bibitem[Daniel et~al.(1990)]{daniel1990applied}
Wayne~W Daniel et~al.
\newblock Applied nonparametric statistics.
\newblock 1990.

\bibitem[Devlin et~al.(2019)Devlin, Chang, Lee, and Toutanova]{devlin2019bert}
Jacob Devlin, Ming-Wei Chang, Kenton Lee, and Kristina Toutanova.
\newblock Bert: Pre-training of deep bidirectional transformers for language
  understanding, 2019.

\bibitem[Dong \& Yang(2020)Dong and Yang]{dong2020bench}
Xuanyi Dong and Yi~Yang.
\newblock Nas-bench-201: Extending the scope of reproducible neural
  architecture search.
\newblock \emph{arXiv preprint arXiv:2001.00326}, 2020.

\bibitem[Du et~al.(2018)Du, Lee, Tian, Singh, and Poczos]{du2018gradient}
Simon Du, Jason Lee, Yuandong Tian, Aarti Singh, and Barnabas Poczos.
\newblock Gradient descent learns one-hidden-layer cnn: Don’t be afraid of
  spurious local minima.
\newblock In \emph{International Conference on Machine Learning}, pp.\
  1339--1348. PMLR, 2018.

\bibitem[Falkner et~al.(2018)Falkner, Klein, and Hutter]{falkner2018bohb}
Stefan Falkner, Aaron Klein, and Frank Hutter.
\newblock Bohb: Robust and efficient hyperparameter optimization at scale.
\newblock In \emph{International Conference on Machine Learning}, pp.\
  1437--1446. PMLR, 2018.

\bibitem[Frankle \& Carbin(2018)Frankle and Carbin]{frankle2018lottery}
Jonathan Frankle and Michael Carbin.
\newblock The lottery ticket hypothesis: Finding sparse, trainable neural
  networks.
\newblock \emph{arXiv preprint arXiv:1803.03635}, 2018.

\bibitem[Ge et~al.(2017)Ge, Lee, and Ma]{ge2017learning}
Rong Ge, Jason~D Lee, and Tengyu Ma.
\newblock Learning one-hidden-layer neural networks with landscape design.
\newblock \emph{arXiv preprint arXiv:1711.00501}, 2017.

\bibitem[Goodfellow et~al.(2014)Goodfellow, Vinyals, and
  Saxe]{goodfellow2014qualitatively}
Ian~J Goodfellow, Oriol Vinyals, and Andrew~M Saxe.
\newblock Qualitatively characterizing neural network optimization problems.
\newblock \emph{arXiv preprint arXiv:1412.6544}, 2014.

\bibitem[He et~al.(2015)He, Zhang, Ren, and Sun]{he2015deep}
Kaiming He, Xiangyu Zhang, Shaoqing Ren, and Jian Sun.
\newblock Deep residual learning for image recognition, 2015.

\bibitem[He et~al.(2021)He, Zhao, and Chu]{he2021automl}
Xin He, Kaiyong Zhao, and Xiaowen Chu.
\newblock Automl: A survey of the state-of-the-art.
\newblock \emph{Knowledge-Based Systems}, 212:\penalty0 106622, 2021.

\bibitem[Howard et~al.(2019)Howard, Sandler, Chu, Chen, Chen, Tan, Wang, Zhu,
  Pang, Vasudevan, Le, and Adam]{Howard_2019_ICCV}
Andrew Howard, Mark Sandler, Grace Chu, Liang-Chieh Chen, Bo~Chen, Mingxing
  Tan, Weijun Wang, Yukun Zhu, Ruoming Pang, Vijay Vasudevan, Quoc~V. Le, and
  Hartwig Adam.
\newblock Searching for mobilenetv3.
\newblock In \emph{Proceedings of the IEEE/CVF International Conference on
  Computer Vision (ICCV)}, October 2019.

\bibitem[Ioffe \& Szegedy(2015)Ioffe and Szegedy]{ioffe2015batch}
Sergey Ioffe and Christian Szegedy.
\newblock Batch normalization: Accelerating deep network training by reducing
  internal covariate shift, 2015.

\bibitem[Jacot et~al.(2018)Jacot, Gabriel, and Hongler]{jacot2018neural}
Arthur Jacot, Franck Gabriel, and Cl{\'e}ment Hongler.
\newblock Neural tangent kernel: Convergence and generalization in neural
  networks.
\newblock \emph{arXiv preprint arXiv:1806.07572}, 2018.

\bibitem[Kingma \& Ba(2017)Kingma and Ba]{kingma2017adam}
Diederik~P. Kingma and Jimmy Ba.
\newblock Adam: A method for stochastic optimization, 2017.

\bibitem[Krizhevsky(2009)]{Krizhevsky09learningmultiple}
Alex Krizhevsky.
\newblock Learning multiple layers of features from tiny images.
\newblock Technical report, 2009.

\bibitem[Krizhevsky et~al.(2017)Krizhevsky, Sutskever, and
  Hinton]{Krizhevsky_2017}
Alex Krizhevsky, Ilya Sutskever, and Geoffrey~E. Hinton.
\newblock Imagenet classification with deep convolutional neural networks.
\newblock \emph{Communications of the ACM}, 60\penalty0 (6):\penalty0 84–90,
  5 2017.
\newblock ISSN 1557-7317.
\newblock \doi{10.1145/3065386}.
\newblock URL \url{http://dx.doi.org/10.1145/3065386}.

\bibitem[Lee et~al.(2018)Lee, Ajanthan, and Torr]{lee2018snip}
Namhoon Lee, Thalaiyasingam Ajanthan, and Philip~HS Torr.
\newblock Snip: Single-shot network pruning based on connection sensitivity.
\newblock \emph{arXiv preprint arXiv:1810.02340}, 2018.

\bibitem[Li et~al.(2017)Li, Jamieson, DeSalvo, Rostamizadeh, and
  Talwalkar]{li2017hyperband}
Lisha Li, Kevin Jamieson, Giulia DeSalvo, Afshin Rostamizadeh, and Ameet
  Talwalkar.
\newblock Hyperband: A novel bandit-based approach to hyperparameter
  optimization.
\newblock \emph{The Journal of Machine Learning Research}, 18\penalty0
  (1):\penalty0 6765--6816, 2017.

\bibitem[Li \& Yuan(2017)Li and Yuan]{li2017convergence}
Yuanzhi Li and Yang Yuan.
\newblock Convergence analysis of two-layer neural networks with relu
  activation.
\newblock \emph{arXiv preprint arXiv:1705.09886}, 2017.

\bibitem[Lin et~al.(2021)Lin, Wang, Sun, Chen, Sun, Qian, Li, and
  Jin]{lin2021zen}
Ming Lin, Pichao Wang, Zhenhong Sun, Hesen Chen, Xiuyu Sun, Qi~Qian, Hao Li,
  and Rong Jin.
\newblock Zen-nas: A zero-shot nas for high-performance image recognition.
\newblock In \emph{Proceedings of the IEEE/CVF International Conference on
  Computer Vision}, pp.\  347--356, 2021.

\bibitem[Liu et~al.(2018{\natexlab{a}})Liu, Zoph, Neumann, Shlens, Hua, Li,
  Fei-Fei, Yuille, Huang, and Murphy]{liu2018progressive}
Chenxi Liu, Barret Zoph, Maxim Neumann, Jonathon Shlens, Wei Hua, Li-Jia Li,
  Li~Fei-Fei, Alan Yuille, Jonathan Huang, and Kevin Murphy.
\newblock Progressive neural architecture search.
\newblock In \emph{Proceedings of the European conference on computer vision
  (ECCV)}, pp.\  19--34, 2018{\natexlab{a}}.

\bibitem[Liu et~al.(2018{\natexlab{b}})Liu, Simonyan, and Yang]{liu2018darts}
Hanxiao Liu, Karen Simonyan, and Yiming Yang.
\newblock Darts: Differentiable architecture search.
\newblock \emph{arXiv preprint arXiv:1806.09055}, 2018{\natexlab{b}}.

\bibitem[Luo et~al.(2020)Luo, Tan, Wang, Qin, Chen, and Liu]{luo2020semi}
Renqian Luo, Xu~Tan, Rui Wang, Tao Qin, Enhong Chen, and Tie-Yan Liu.
\newblock Semi-supervised neural architecture search.
\newblock \emph{arXiv preprint arXiv:2002.10389}, 2020.

\bibitem[Martens(2014)]{martens2014new}
James Martens.
\newblock New insights and perspectives on the natural gradient method.
\newblock \emph{arXiv preprint arXiv:1412.1193}, 2014.

\bibitem[Mellor et~al.(2021)Mellor, Turner, Storkey, and
  Crowley]{mellor2021neural}
Joseph Mellor, Jack Turner, Amos Storkey, and Elliot~J. Crowley.
\newblock Neural architecture search without training, 2021.

\bibitem[Ning et~al.(2020)Ning, Zheng, Zhao, Wang, and Yang]{ning2020generic}
Xuefei Ning, Yin Zheng, Tianchen Zhao, Yu~Wang, and Huazhong Yang.
\newblock A generic graph-based neural architecture encoding scheme for
  predictor-based nas.
\newblock In \emph{Computer Vision--ECCV 2020: 16th European Conference,
  Glasgow, UK, August 23--28, 2020, Proceedings, Part XIII 16}, pp.\  189--204.
  Springer, 2020.

\bibitem[Park et~al.(2020)Park, Lee, Peng, Cao, and
  Sohl-Dickstein]{park2020towards}
Daniel~S Park, Jaehoon Lee, Daiyi Peng, Yuan Cao, and Jascha Sohl-Dickstein.
\newblock Towards nngp-guided neural architecture search.
\newblock \emph{arXiv preprint arXiv:2011.06006}, 2020.

\bibitem[Paszke et~al.(2017)Paszke, Gross, Chintala, Chanan, Yang, DeVito, Lin,
  Desmaison, Antiga, and Lerer]{paszke2017automatic}
Adam Paszke, Sam Gross, Soumith Chintala, Gregory Chanan, Edward Yang, Zachary
  DeVito, Zeming Lin, Alban Desmaison, Luca Antiga, and Adam Lerer.
\newblock Automatic differentiation in pytorch.
\newblock 2017.

\bibitem[Pham et~al.(2018)Pham, Guan, Zoph, Le, and Dean]{pham2018efficient}
Hieu Pham, Melody Guan, Barret Zoph, Quoc Le, and Jeff Dean.
\newblock Efficient neural architecture search via parameters sharing.
\newblock In \emph{International Conference on Machine Learning}, pp.\
  4095--4104. PMLR, 2018.

\bibitem[Radosavovic et~al.(2019)Radosavovic, Johnson, Xie, Lo, and
  Doll{\'a}r]{radosavovic2019network}
Ilija Radosavovic, Justin Johnson, Saining Xie, Wan-Yen Lo, and Piotr
  Doll{\'a}r.
\newblock On network design spaces for visual recognition.
\newblock In \emph{Proceedings of the IEEE/CVF International Conference on
  Computer Vision}, pp.\  1882--1890, 2019.

\bibitem[Real et~al.(2019)Real, Aggarwal, Huang, and Le]{real2019regularized}
Esteban Real, Alok Aggarwal, Yanping Huang, and Quoc~V Le.
\newblock Regularized evolution for image classifier architecture search.
\newblock In \emph{Proceedings of the aaai conference on artificial
  intelligence}, volume~33, pp.\  4780--4789, 2019.

\bibitem[Rumelhart et~al.(1986)Rumelhart, Hinton, and
  Williams]{rumelhart1986learning}
David~E Rumelhart, Geoffrey~E Hinton, and Ronald~J Williams.
\newblock Learning representations by back-propagating errors.
\newblock \emph{nature}, 323\penalty0 (6088):\penalty0 533--536, 1986.

\bibitem[Russakovsky et~al.(2015)Russakovsky, Deng, Su, Krause, Satheesh, Ma,
  Huang, Karpathy, Khosla, Bernstein, et~al.]{russakovsky2015imagenet}
Olga Russakovsky, Jia Deng, Hao Su, Jonathan Krause, Sanjeev Satheesh, Sean Ma,
  Zhiheng Huang, Andrej Karpathy, Aditya Khosla, Michael Bernstein, et~al.
\newblock Imagenet large scale visual recognition challenge.
\newblock \emph{International journal of computer vision}, 115\penalty0
  (3):\penalty0 211--252, 2015.

\bibitem[Serra et~al.(2018)Serra, Tjandraatmadja, and
  Ramalingam]{serra2018bounding}
Thiago Serra, Christian Tjandraatmadja, and Srikumar Ramalingam.
\newblock Bounding and counting linear regions of deep neural networks.
\newblock In \emph{International Conference on Machine Learning}, pp.\
  4558--4566. PMLR, 2018.

\bibitem[Siems et~al.(2020)Siems, Zimmer, Zela, Lukasik, Keuper, and
  Hutter]{siems2020bench}
Julien Siems, Lucas Zimmer, Arber Zela, Jovita Lukasik, Margret Keuper, and
  Frank Hutter.
\newblock Nas-bench-301 and the case for surrogate benchmarks for neural
  architecture search.
\newblock \emph{arXiv preprint arXiv:2008.09777}, 2020.

\bibitem[Soltanolkotabi(2017)]{soltanolkotabi2017learning}
Mahdi Soltanolkotabi.
\newblock Learning relus via gradient descent.
\newblock \emph{arXiv preprint arXiv:1705.04591}, 2017.

\bibitem[Srivastava et~al.(2014)Srivastava, Hinton, Krizhevsky, Sutskever, and
  Salakhutdinov]{JMLR:v15:srivastava14a}
Nitish Srivastava, Geoffrey Hinton, Alex Krizhevsky, Ilya Sutskever, and Ruslan
  Salakhutdinov.
\newblock Dropout: A simple way to prevent neural networks from overfitting.
\newblock \emph{Journal of Machine Learning Research}, 15\penalty0
  (56):\penalty0 1929--1958, 2014.
\newblock URL \url{http://jmlr.org/papers/v15/srivastava14a.html}.

\bibitem[Tan \& Le(2019)Tan and Le]{pmlr-v97-tan19a}
Mingxing Tan and Quoc Le.
\newblock {E}fficient{N}et: Rethinking model scaling for convolutional neural
  networks.
\newblock In Kamalika Chaudhuri and Ruslan Salakhutdinov (eds.),
  \emph{Proceedings of the 36th International Conference on Machine Learning},
  volume~97 of \emph{Proceedings of Machine Learning Research}, pp.\
  6105--6114. PMLR, 09--15 Jun 2019.
\newblock URL \url{https://proceedings.mlr.press/v97/tan19a.html}.

\bibitem[Tanaka et~al.(2020)Tanaka, Kunin, Yamins, and
  Ganguli]{tanaka2020pruning}
Hidenori Tanaka, Daniel Kunin, Daniel~LK Yamins, and Surya Ganguli.
\newblock Pruning neural networks without any data by iteratively conserving
  synaptic flow.
\newblock \emph{arXiv preprint arXiv:2006.05467}, 2020.

\bibitem[Theis et~al.(2018)Theis, Korshunova, Tejani, and
  Husz{\'a}r]{theis2018faster}
Lucas Theis, Iryna Korshunova, Alykhan Tejani, and Ferenc Husz{\'a}r.
\newblock Faster gaze prediction with dense networks and fisher pruning.
\newblock \emph{arXiv preprint arXiv:1801.05787}, 2018.

\bibitem[Turner et~al.(2019)Turner, Crowley, O'Boyle, Storkey, and
  Gray]{turner2019blockswap}
Jack Turner, Elliot~J Crowley, Michael O'Boyle, Amos Storkey, and Gavin Gray.
\newblock Blockswap: Fisher-guided block substitution for network compression
  on a budget.
\newblock \emph{arXiv preprint arXiv:1906.04113}, 2019.

\bibitem[Turner et~al.(2021)Turner, Crowley, and O'Boyle]{turner2021neural}
Jack Turner, Elliot~J Crowley, and Michael~FP O'Boyle.
\newblock Neural architecture search as program transformation exploration.
\newblock In \emph{Proceedings of the 26th ACM International Conference on
  Architectural Support for Programming Languages and Operating Systems}, pp.\
  915--927, 2021.

\bibitem[Wang et~al.(2020)Wang, Zhang, and Grosse]{wang2020picking}
Chaoqi Wang, Guodong Zhang, and Roger Grosse.
\newblock Picking winning tickets before training by preserving gradient flow.
\newblock \emph{arXiv preprint arXiv:2002.07376}, 2020.

\bibitem[Wang et~al.(2019)Wang, Zhao, Jinnai, Tian, and
  Fonseca]{wang2019alphax}
Linnan Wang, Yiyang Zhao, Yuu Jinnai, Yuandong Tian, and Rodrigo Fonseca.
\newblock Alphax: exploring neural architectures with deep neural networks and
  monte carlo tree search.
\newblock \emph{arXiv preprint arXiv:1903.11059}, 2019.

\bibitem[Wen et~al.(2020)Wen, Liu, Chen, Li, Bender, and
  Kindermans]{wen2020neural}
Wei Wen, Hanxiao Liu, Yiran Chen, Hai Li, Gabriel Bender, and Pieter-Jan
  Kindermans.
\newblock Neural predictor for neural architecture search.
\newblock In \emph{European Conference on Computer Vision}, pp.\  660--676.
  Springer, 2020.

\bibitem[Williams(1992)]{williams1992simple}
Ronald~J Williams.
\newblock Simple statistical gradient-following algorithms for connectionist
  reinforcement learning.
\newblock \emph{Machine learning}, 8\penalty0 (3):\penalty0 229--256, 1992.

\bibitem[Xiao et~al.(2020)Xiao, Pennington, and
  Schoenholz]{xiao2020disentangling}
Lechao Xiao, Jeffrey Pennington, and Samuel Schoenholz.
\newblock Disentangling trainability and generalization in deep neural
  networks.
\newblock In \emph{International Conference on Machine Learning}, pp.\
  10462--10472. PMLR, 2020.

\bibitem[Ying et~al.(2019)Ying, Klein, Christiansen, Real, Murphy, and
  Hutter]{ying2019bench}
Chris Ying, Aaron Klein, Eric Christiansen, Esteban Real, Kevin Murphy, and
  Frank Hutter.
\newblock Nas-bench-101: Towards reproducible neural architecture search.
\newblock In \emph{International Conference on Machine Learning}, pp.\
  7105--7114. PMLR, 2019.

\bibitem[Zhang et~al.(2016)Zhang, Bengio, Hardt, Recht, and
  Vinyals]{zhang2016understanding}
Chiyuan Zhang, Samy Bengio, Moritz Hardt, Benjamin Recht, and Oriol Vinyals.
\newblock Understanding deep learning requires rethinking generalization.
\newblock \emph{arXiv preprint arXiv:1611.03530}, 2016.

\bibitem[Zhou et~al.(2020)Zhou, Zhou, Zhang, Loy, Yi, Zhang, and
  Ouyang]{zhou2020econas}
Dongzhan Zhou, Xinchi Zhou, Wenwei Zhang, Chen~Change Loy, Shuai Yi, Xuesen
  Zhang, and Wanli Ouyang.
\newblock Econas: Finding proxies for economical neural architecture search.
\newblock In \emph{Proceedings of the IEEE/CVF Conference on Computer Vision
  and Pattern Recognition}, pp.\  11396--11404, 2020.

\bibitem[Zoph \& Le(2016)Zoph and Le]{zoph2016neural}
Barret Zoph and Quoc~V Le.
\newblock Neural architecture search with reinforcement learning.
\newblock \emph{arXiv preprint arXiv:1611.01578}, 2016.

\bibitem[Zoph et~al.(2018)Zoph, Vasudevan, Shlens, and Le]{zoph2018learning}
Barret Zoph, Vijay Vasudevan, Jonathon Shlens, and Quoc~V Le.
\newblock Learning transferable architectures for scalable image recognition.
\newblock In \emph{Proceedings of the IEEE conference on computer vision and
  pattern recognition}, pp.\  8697--8710, 2018.

\end{thebibliography}
\bibliographystyle{iclr2022_conference}

\appendix
\clearpage


\appendix

\section{Appendix}
\label{sec:appendix}
\subsection{Figure}

\begin{figure}[H]
     \centering
     \begin{subfigure}[b]{0.3\textwidth}
         \centering
         \includegraphics[width=\textwidth]{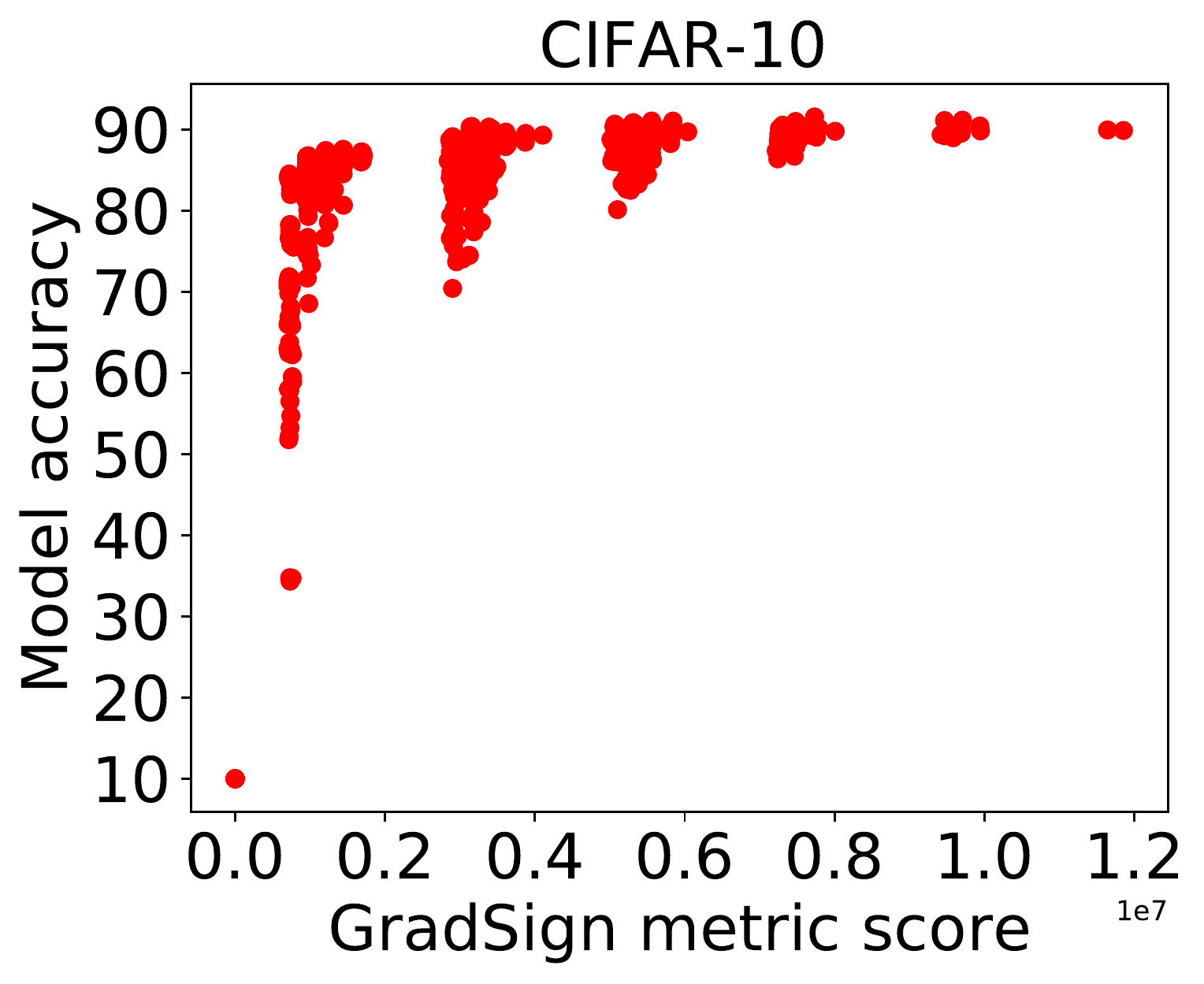}
         \label{fig:201_CIFAR10}
     \end{subfigure}
     \hfill
     \begin{subfigure}[b]{0.3\textwidth}
         \centering
         \includegraphics[width=\textwidth]{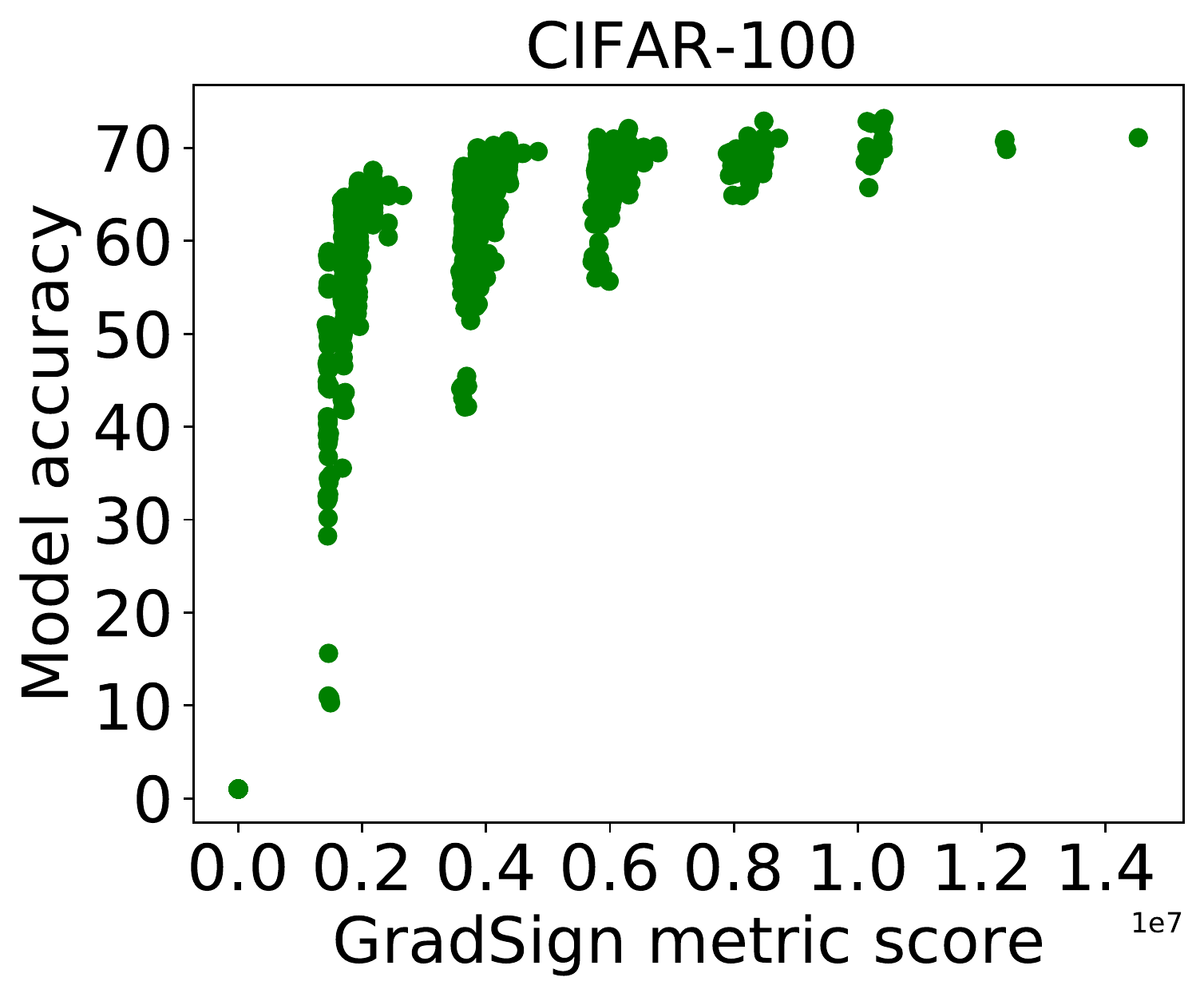}
         \label{fig:201_CIFAR100}
     \end{subfigure}
     \hfill
     \begin{subfigure}[b]{0.3\textwidth}
         \centering
         \includegraphics[width=\textwidth]{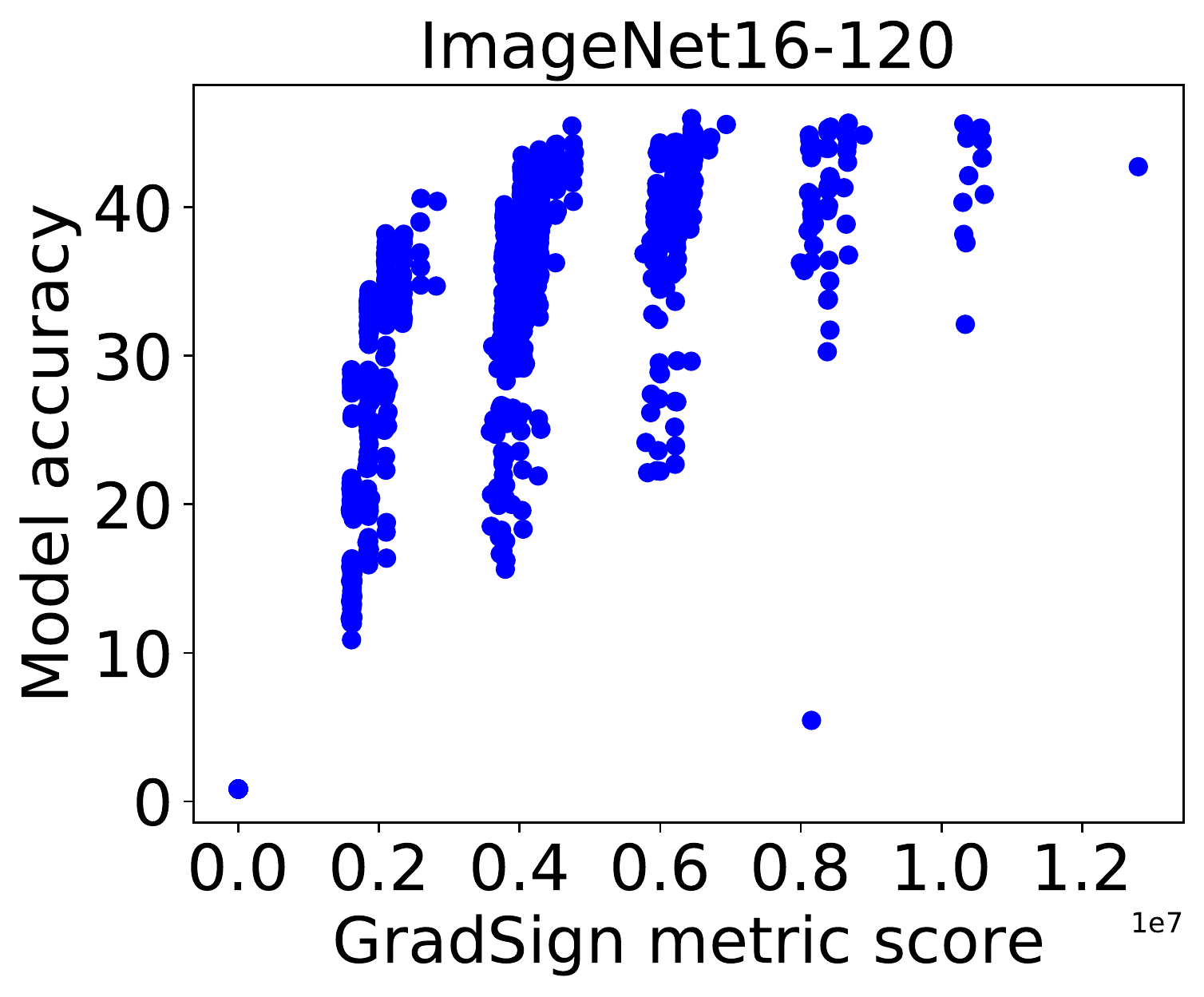}
         \label{fig:201_IN}
     \end{subfigure}
        \caption{Visualization of model testing accuracy versus GradSign metric score on CIFAR10, CIFAR100, ImageNet16-120.}
        \label{fig:201}
\end{figure}

\subsection{Proof}
\label{sec:app-proof}

\noindent\textbf{Lemma 1 Proof:} \label{proof:lemma_1}
For a single training sample $(x_i, y_i)$, we minimize its objective function $l(f_\theta(x_i), y_i)$ with gradient descent: 
\begin{eqnarray}
\nabla_\theta l(f_\theta(x_i), y_i) &=& \frac{\partial l(f_\theta(x_i), y_i)}{\partial f_\theta(x_i)}\cdot f^\prime_\theta(x_i)
\end{eqnarray}
At any local optimal point $\theta_i^*$, we have $\nabla_\theta l(f_\theta(x_i), y_i)|_{\theta_i^*}=\mathbf{0}$, but $f^\prime_\theta(x_i) \neq \mathbf{0}$ for conventional neural architectures with at least one dense layer~\footnote{The gradient values corresponding to the bias term of the last dense layer are always non-zero.}. Therefore, we show $\frac{\partial l(f_\theta(x_i), y_i)}{\partial f_\theta(x_i)}|_{\theta_i^*}$ must be equal to $0$.
For the commonly used objective functions, such as Mean Squared Error Loss and Cross Entropy Loss, this derivative is equal to $C(f_\theta(x_i) - y_i)$, where $C$ is a non-zero constant.
Hence we have $f_{\theta_i^*}(x_i) = y_i$ and $l(f_{\theta_i^*}(x_i), y_i)=0$ at local optima $\theta_i^*$, which makes $\theta_i^*$ also a global optima as it is impossible to obtain a lower loss value for this single sample. In addition, at local optima $\theta_i^*$, we have:
\begin{eqnarray}
\nabla^2_\theta l(f_\theta(x_i), y_i) &=& \nabla_\theta (\frac{\partial l(f_\theta(x_i), y_i)}{\partial f_\theta(x_i)}\cdot f^\prime_\theta(x_i))\\
&=&\frac{\partial^2 l(f_\theta(x_i), y_i)}{\partial f_\theta(x_i)^2}\cdot f^\prime_\theta(x_i)f^\prime_\theta(x_i)^\top + \frac{\partial l(f_\theta(x_i), y_i)}{\partial f_\theta(x_i)}\cdot f^{\prime\prime}_\theta(x_i)\\
&=& C\cdot f^\prime_\theta(x_i)f^\prime_\theta(x_i)^\top
\end{eqnarray}
The above equation implies that $\nabla^2_\theta l(f_\theta(x_i), y_i)$ is a positive semi-definite matrix, since $C>0$.
This concludes the proof of non existence of saddle points in a sample-wise optimization landscape. This result aligns with our convexity assumptions in Section~\ref{sec:notation}.\\

\noindent\textbf{Theorem 2 Proof:} \label{proof:claim_1}
Recall that $\theta^*$ denotes a local optima of $J$ which could be reached by a gradient-flow based optimizer start from $\theta_0$.
Since $\theta_0$ is randomly sampled and $[\nabla^2 l(f_\theta(x_i), y_i)]_{k,k} \leq \mathcal{H}$, we have:
\begin{eqnarray}
\label{eq:c2_1}
J &=& \frac{1}{n} \sum_{i}l(f_{\theta^*}(x_i), y_i)\\
\label{eq:c2_2}
&\leq& \frac{1}{n} \sum_{i}\mathcal{H} \cdot \|\theta^*-\theta_i^*\|_2^2\\
\label{eq:c2_3}
&\leq& \frac{\mathcal{H}}{n}\sum_i \|\theta^*-\theta_i^*\|_1^2\\
\label{eq:c2_4}
&\leq& \frac{\mathcal{H}}{n}\sum_{i, j} \|\theta_i^*-\theta_j^*\|_1^2\\
\label{eq:c2_5}
&\leq& n^3 \Psi_{\mathcal{S}, l}^2(f_{\theta_0}(\cdot))
\end{eqnarray}\\
where Eq~\ref{eq:c2_1} $\rightarrow$ Eq~\ref{eq:c2_2} uses the basic property of the smoothness upper bound $\mathcal{H}$ and the fact that each local optimum $\theta_i^*$ satisfies $l(f_{\theta_i^*}(x_i), y_i)=0$. Eq~\ref{eq:c2_2} $\rightarrow$ Eq~\ref{eq:c2_3} uses Jensen Inequality for square root operators. Eq~\ref{eq:c2_3} $\rightarrow$ Eq~\ref{eq:c2_4} is derived from the fact that for each dimension in $(\theta^*-\theta_i^*)$ we have:
\begin{eqnarray}
|[\theta^*-\theta_i^*]_k| &\leq& \sum_{j}|[\theta_j^*-\theta_i^*]_k|
\end{eqnarray}
Otherwise, $\theta^*$ dose not lie in the convex hull of $\{\theta_i^*\}_{i\in[n]}$ which contradicts with our assumption stated in Section~\ref{sec:notation}.
The bound is tight when $\theta_i^*=\theta_j^*, \forall i,j \in [n]$. 

Eq~\ref{eq:c2_1}$\rightarrow$Eq~\ref{eq:c2_2}: As we have $\nabla^2 l(f_\theta(x_i), y_i) \preceq \mathcal{H}\mathbb{I}$ and $l(f_{\theta_i^*}(x_i), y_i)=0, \nabla_\theta l(f_\theta(x_i), y_i)|_{\theta_i^*}=\mathbf{0}$, we could derive the following inequality:
\begin{eqnarray}
l(f_{\theta^*}(x_i), y_i) &\leq& l(f_{\theta_i^*}(x_i), y_i) + \nabla_\theta l(f_{\theta_i^*}(x_i), y_i)^\top (\theta^*-\theta_i^*) +  \frac{\mathcal{H}}{2}\|\theta^*-\theta_i^*\|_2^2\\
&=& \frac{\mathcal{H}}{2} \|\theta^*-\theta_i^*\|_2^2
\end{eqnarray}
we thus have $\frac{1}{n} \sum_{i}l(f_{\theta^*}(x_i), y_i) \leq \frac{1}{2n} \sum_{i}\mathcal{H} \cdot \|\theta^*-\theta_i^*\|_2^2$.
\\

\noindent\textbf{Theorem 3 Proof:}\label{proof:claim_2}
Let $\mathbb{E}_{(x_u, y_u)\sim \mathcal{D}}[l(f_\theta^*(x_u), y_u)]$ denotes the true population error. With probability $1-\delta$, we have:
\begin{eqnarray}
\label{eq:c3_1}
\mathbb{E}_{(x_u, y_u)\sim \mathcal{D}}[l(f_\theta^*(x_u), y_u)] &\leq& \mathcal{H} \mathbb{E}_{(x_u, y_u)\sim \mathcal{D}}[\|\theta^*-\theta_u^*\|_1^2]\\
\label{eq:c3_2}
&\leq& \frac{\mathcal{H}}{n}\sum_{i=1}^n\|\theta^*-\theta_i^*\|^2_1 + \frac{\sigma}{\sqrt{n\delta}}\\
\label{eq:c3_3}
&\leq& n^3 \Psi_{\mathcal{S}, l}^2(f_{\theta_0}(\cdot)) + \frac{\sigma}{\sqrt{n\delta}}
\end{eqnarray}
where $n$ and $\sigma$ are constants, $\mathcal{S}$ is a training dataset, and $\mathcal{D}$ denotes its underlying data distribution.
This implies that $\Psi_{\mathcal{S}, l}(f_{\theta_0}(\cdot))$ is an accurate indicator for the true population loss. Eq~\ref{eq:c3_1} $\rightarrow$ Eq~\ref{eq:c3_2} uses Chebyshev's inequality, while Eq~\ref{eq:c3_2} $\rightarrow$ Eq~\ref{eq:c3_3} uses the same inequality derived in \textbf{Claim 2}.

\noindent\textbf{Algorithm Proof:}
Given that sample-wise local optima $\{[\theta_i^*], i\in[n]\}$ are contained in the convex area around $\theta_0$. We derive the following property:

\begin{eqnarray}
\text{sign}([\theta_i^*-\theta_0]_{k}) &=& \text{sign}([\nabla_\theta l(f_\theta(x_i), y_i)|_{\theta_0}]_{k})
\end{eqnarray}

 Since $\theta_0$ is a randomly chosen initialization point, without loss of generality, we assume $\theta_0$ is sampled from a hypercube $[-a, a]$. Thus we have:
 
\begin{eqnarray}
P(\text{sign}([\nabla_\theta l(f_\theta(x_i), y_i)|_{\theta_0}]_{k})\neq\text{sign}([\nabla_\theta l(f_\theta(x_j), y_j)|_{\theta_0}]_{k})) &=& \frac{|[\theta_i^*]_{k}-[\theta_j^*]_{k}|}{2a}
\end{eqnarray}

Where $P(\text{sign}([\nabla_\theta l(f_\theta(x_i), y_i)|_{\theta_0}]_{k})\neq\text{sign}([\nabla_\theta l(f_\theta(x_j), y_j)|_{\theta_0}]_{k}))$ denotes the probability for $[\nabla_\theta l(f_\theta(x_i), y_i)|_{\theta_0}]_{k}$ and $[\nabla_\theta l(f_\theta(x_j), y_j)|_{\theta_0}]_{k}$ having different signs. Notice that we have completely dropped the dependency for $\theta_i^*$ at this point and can simply infer from $\text{sign}([\nabla_\theta l(f_\theta(x_i), y_i)|_{\theta_0}]_{k})$. To complete our proof:

\begin{eqnarray}
\Psi_{\mathcal{S}, l}(f_{\theta_0}(\cdot))&=&\frac{\sqrt{\mathcal{H}}}{n}\sum_{i, j}\|\theta_i^*-\theta_j^*\|_1\\
&=&\frac{2a\sqrt{\mathcal{H}}}{n}\sum_{k}\sum_{i,j}P(\text{sign}([\nabla_\theta l(f_\theta(x_i), y_i)|_{\theta_0}]_{k})\neq\text{sign}([\nabla_\theta l(f_\theta(x_j), y_j)|_{\theta_0}]_{k}))\\
\label{eq:final}
&\propto& n^2-\sum_{k}\sum_{i,j}P(\text{sign}([\nabla_\theta l(f_\theta(x_i), y_i)|_{\theta_0}]_{k})=\text{sign}([\nabla_\theta l(f_\theta(x_j), y_j)|_{\theta_0}]_{k}))
\end{eqnarray}

\noindent\textbf{Simplification Proof:}: for each $i,j$ and a given $k$, we could estimate $P(\text{sign}([\nabla_\theta l(f_\theta(x_i), y_i)|_{\theta_0}]_{k})=\text{sign}([\nabla_\theta l(f_\theta(x_j), y_j)|_{\theta_0}]_{k}))$ using:

\begin{eqnarray}
\frac{|\text{sign}([\nabla_\theta l(f_\theta(x_i), y_i)|_{\theta_0}]_{k})+\text{sign}([\nabla_\theta l(f_\theta(x_j), y_j)|_{\theta_0}]_{k})|}{2}
\label{eq:sim}
\end{eqnarray}

which is valid (not equal to zero) only when $\text{sign}([\nabla_\theta l(f_\theta(x_i), y_i)|_{\theta_0}]_{k})==\text{sign}([\nabla_\theta l(f_\theta(x_j), y_j)|_{\theta_0}]_{k})$. Suppose we have p positive and n-p negative $\text{sign}([\nabla_\theta l(f_\theta(x_i), y_i)|_{\theta_0}]_{k})$ for n samples, since we only care about samples share the same sign, the original probability estimation is $\frac{(n-p)^2+p^2}{n^2}=\frac{1}{2} + \frac{(n-2p)^2}{2n^2}$. Thus we only need to measure the quantity $|n-2p|$ which simply equals to $|\sum_i \text{sign}([\nabla_\theta l(f_\theta(x_i), y_i)|_{\theta_0}]_{k})|$. 

\subsection{Experiments Setup}
\label{app_exp}
The code we used during experimentation was mainly based on existed code base \cite{abdelfattah2021zerocost, mellor2021neural}\cite{abdelfattah2021zerocost} which is under Apache-2.0 License.

The hardwares we used were Amazon EC2 C5 instances with no GPU involved and p3 instance with one V100 Tensor Core GPU.

As our methods is gradient-based which is training free, we don't need to split our dataset. For Spearman's correlation measurement on NAS-Bench-201, we set batch size to 64, which is used by most baselines. For the Kendall's Tau experiment, other accuracy comparison experience and the GradSign assisted algorithms, we used a batch size of 128, also to match the batch size in other baselines (NASWOT). We use Pytorch default parameter initialization for all architectures. Random seed in correlation experiments is set to 42 which is also randomly chosen. For accuracy experiments, our results are summarized over 500 runs whose random seed are chosen randomly for each run. For the correlation evaluation of each individual architecture, we only use one $\theta_0$ for minimizing computational cost. Our approach can be easily generalized to an average of multiple $\theta_0$s and can trade-off between efficiency and accuracy.

\subsection{Additional results}

\begin{figure}[H]
\begin{center}
\includegraphics[width=0.65\textwidth]{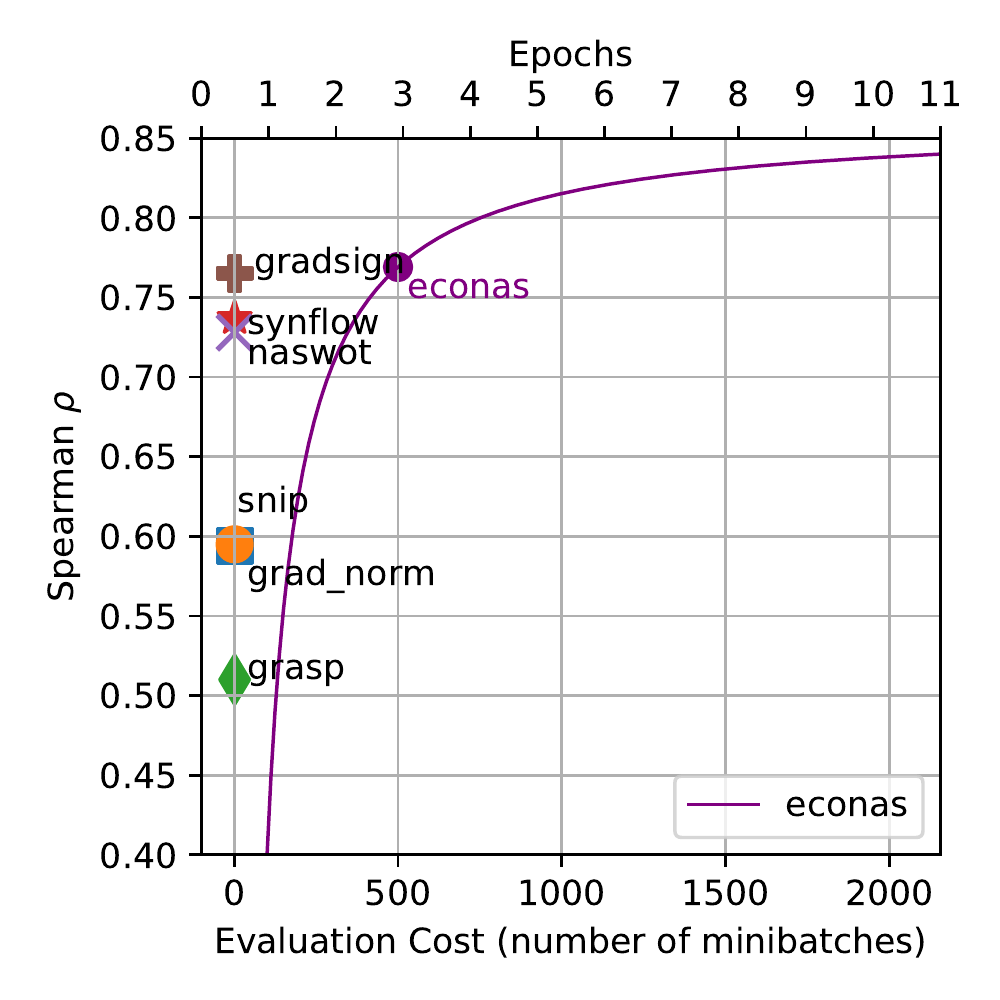}
\caption{Comparison with sample-based methods (EconNAS) on NAS-Bench-201 across CIFAR-10. EconNAS requires more than ~500 minibatches to have a better performance than GradSign while gradient-based methods only require 1 minibatch.}
\label{fig:nb2cifar10}
\end{center}
\end{figure}

\noindent\textbf{Sample-based:} Fig~\ref{fig:nb2cifar10} compares EconNAS, a sample-based method, with existing gradient-based methods on MPI. Results of EconNAS are referenced from~\citet{abdelfattah2021zerocost}. To achieve a similar MPI performance as GradSign, EconNAS needs ~500 minibatches of samples for each candidate's proxy training,  while all gradient-based methods (including GradSign) require only one minibatch. By increasing the number of minibatches, EconNAS can achieve higher Spearman's $\rho$ scores, which eventually converge to 0.85. At that point overfitting takes place and the score cannot be further improved.

\noindent\textbf{Learning-based:} Table~\ref{tab:Learning-Based} compares GradSign and existing learning-based methods (MLP, LSTM, and GATES) on the Kendall's Tau correlation score. 
The MLP, LSTM, and GATES results are referenced from \citet{ning2020generic}. 
For MLP and LSTM \citep{wang2019alphax}, the predictor uses Multi Layer Perceptron (MLP) and Long Short Term Memory (LSTM) as the base predictor, while GATES \citep{ning2020generic} uses Graph Neural Network (GNN) as the base predictor. All results are obtained on NAS-Bench-201 and the GradSign's score is averaged over all three datasets (CIFAR-10, CIFAR-100 and ImageNet16-120) as \citep{ning2020generic} does not provide which dataset is used for calculating Kendall's Tau.

To achieve a similar score as GradSign, MLP, LSTM and GATES-1 require an average of 1959, 978 and 1959 minibatches per sample respectively to prepare the dataset for training the predictors. Although GATES-2 achieves a better correlation score than GradSign, it still needs 195 minibatches per sample to prepare the dataset for training the GATES-2 predictor. In addition to the cost of preparing a training dataset,  each predictor also has to be trained on the dataset as well, which involves 200 more epochs while the cost of evaluating GradSign is one mini-batch. With less mini-batches evaluated for learning-based methods, their training set sizes shrink significantly (e.g., 195 mini-batches equal to 78 training samples and 7813 testing samples). This may result in overfitting to the training set.

\begin{table}[H]
\caption{\label{tab:Learning-Based} Comparison with learning-based methods (MLP, LSTM and GATES) on NAS-Bench-201. GATES-1 represents GATES predictor with only one layer and GATES-2 denotes GATES predictors with more than one layers.}
\begin{center}
\begin{tabular}{ccc}
\hline
         & Kendall's Tau & Average minibatches per sample \\ \hline
MLP      & 0.5388        & 1959                           \\
LSTM     & 0.6407        & 978                            \\
GATES-1  & 0.45          & 1959                           \\
GATES-2  & 0.7401        & 195                            \\
\ours{GradSign} & 0.6016        & 1                              \\ \hline
\end{tabular}
\end{center}
\end{table} 

We also includes the evaluation of \algname for both MPI correlation performance and GradSign-assisted NAS algorithms on the latest version of NAS-Bench-201 (NATS-Bench) across three datasets (CIFAR-10, CIFAR-100 and ImageNet16-120) in Table~\ref{tab:NATS-Spearman}, Table~\ref{tab:NATS-Algo}. Results show \algname is robust against hyper-parameter tuning as long as the trained networks can converge to near optimal. Also notice that GradSign-assisted NAS algorithms could not only achieve a better accuracy but lower variance as well compared to their non-assisted counterparts.

\begin{table}[H]
\caption{\label{tab:NATS-Spearman} Spearman's $\rho$ evaluated on the latest version of NAS-Bench-201 (NATS-Bench)}
\begin{center}
\begin{tabular}{lccc}
\hline
              & CIFAR-10 & CIFAR-100 & ImageNet16-120 \\ \hline
NAS-Bench-201 & 0.765    & 0.793     & 0.783          \\
NATS-Bench    & 0.760    & 0.792     & 0.784          \\ \hline
\end{tabular}
\end{center}
\end{table}

\setlength{\tabcolsep}{3pt}
\begin{table}[H]
\caption{\label{tab:NATS-Algo} Mean $\pm$ std accuracy evaluated over NATS-Bench. All results are averaged over 500 runs. To make a fair comparison across all the methods, the search is performed on CIFAR-100 dataset while the architectures' performance are evaluated over CIFAR-10, CIFAR-100 and ImageNet16-120. All the methods have a search time budget of 12000s. Note that the benchmark results might not match with the original paper as we have run all the experiments from start in a environment different from \cite{dong2020bench}.}
\begin{center}
\begin{tabular}{ccccccc}
\hline
      \multirow{2}{*}{Methods}      & \multicolumn{2}{c}{CIFAR-10}                & \multicolumn{2}{c}{CIFAR-100}               & \multicolumn{2}{c}{ImageNet16-120}          \\ \cline{2-7} 
 & Validation           & Test                 & Validation           & Test                 & Validation           & Test                 \\ \hline
REA         & 91.06$\pm$0.49          & 93.84$\pm$0.45          & 71.53$\pm$1.31          & 71.60$\pm$1.27          & 44.82$\pm$1.23          & 45.18$\pm$1.37          \\
\ours{G-REA}       & \textbf{91.35$\pm$0.35} & \textbf{94.15$\pm$0.32} & \textbf{72.67$\pm$1.05} & \textbf{72.65$\pm$0.97} & \textbf{45.55$\pm$0.96} & \textbf{45.99$\pm$0.93}  \\ \hline
RS          & 90.95$\pm$0.28          & 93.77$\pm$0.26          & 71.01$\pm$0.97          & 71.15$\pm$0.95          & 44.58$\pm$0.95          & 44.73$\pm$1.10          \\
\ours{G-RS}        & \textbf{91.23$\pm$0.22} & \textbf{94.02$\pm$0.22} & \textbf{72.12$\pm$0.82} & \textbf{72.15$\pm$0.78} & \textbf{45.43$\pm$0.74} & \textbf{45.83$\pm$0.80} \\ \hline
REINFORCE   & 90.92$\pm$0.38          & 93.71$\pm$0.37          & 71.04$\pm$1.02 & 71.17$\pm$1.12          & 44.56$\pm$0.97          & 44.80$\pm$1.18          \\
\ours{G-REINFORCE} & \textbf{91.20$\pm$0.23} & \textbf{93.98$\pm$0.23} & \textbf{71.93$\pm$0.91}          & \textbf{72.05$\pm$0.89} & \textbf{45.28$\pm$0.77} & \textbf{45.64$\pm$0.86} \\ \hline
\end{tabular}
\end{center}
\end{table}

To demonstrate the potential for \algname in a more complicated computer vision task, we compare the performance of \algname with ZenNAS following their setups and search space.  Due to the limitation of computational resources\footnote{the original setup in ZenNAS could take up to 8 months for training 480 epochs on ImageNet-1k}, we only run 10000 evolution iterations using solely Zen score or \algname score to select the architecture candidate and 20 epochs to train the selected architecture. Following ZenNAS's setup, EfficientNet-B3 is used as teacher network when training selected architectures. Though the top-1 validation accuracy of GradSign in the first 20 epochs is slightly better than Zen, we should note that this process can highly depend on the random seed for evolution search phase. As we mentioned before, ZenNAS uses linear region analysis which makes it less flexible for arbitrary activation functions. On the other hand, since ZenNAS calculates an architecture complexity related score which is both dataset independent and initialization independent, it can be too general and results in low Spearman's $\rho$ as shown in Table~\ref{tab:NAS-201}.

\begin{figure}[H]
\begin{center}
\includegraphics[width=0.7\textwidth]{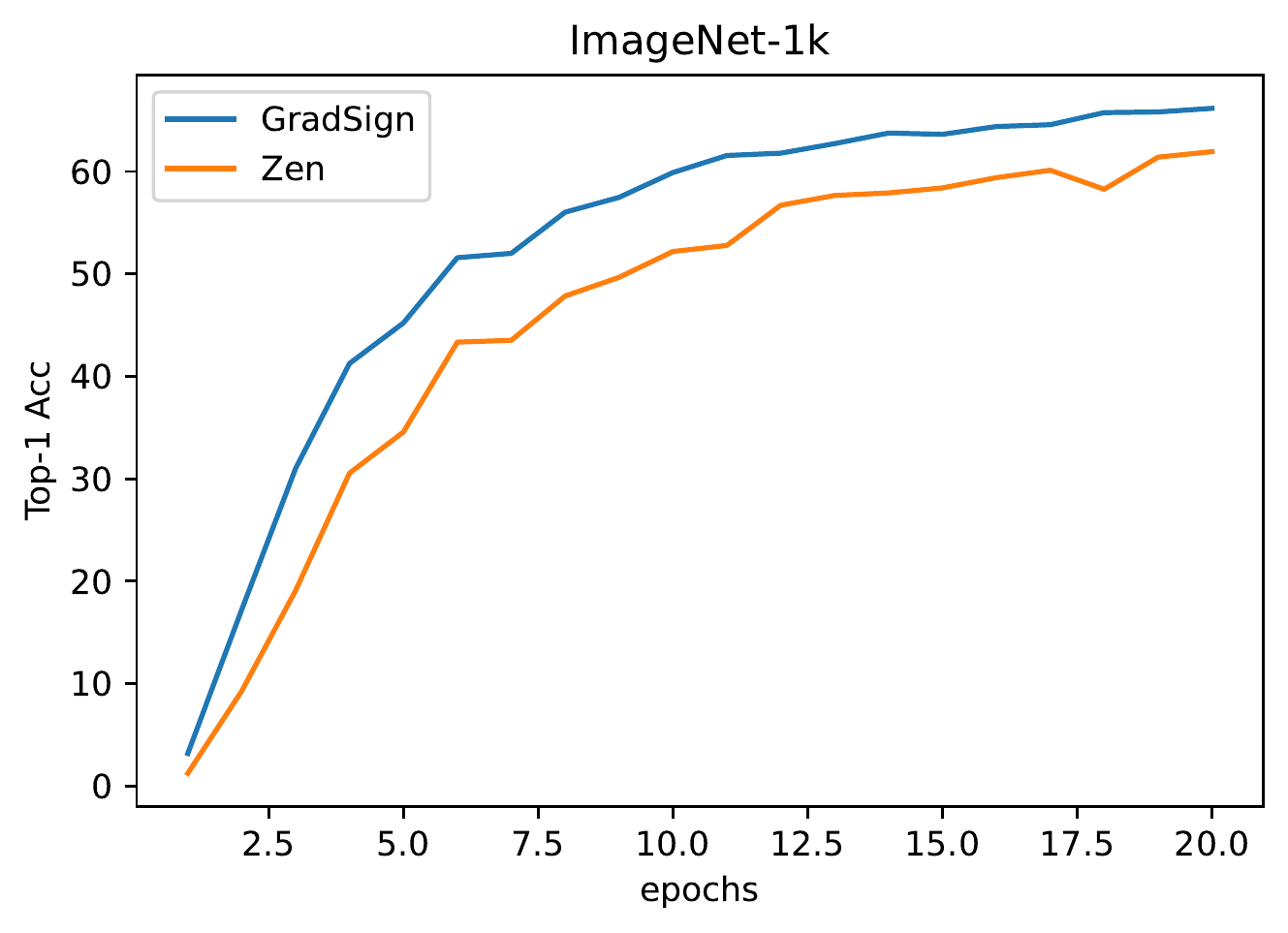}
\caption{Comparison with ZenNAS in their search space on ImageNet-1k. Due to the limitation of computational resources, we only run 10000 evolution iterations and 20 epochs to train the selected architecture. We plot the top-1 prediction accuracy along training for two methods (ZenNAS, GradSign).}
\label{fig:ImageNet1k}
\end{center}
\end{figure}



\subsection{GradSign assisted NAS algorithms}
\begin{algorithm}[]
\SetAlgoLined
\KwResult{Find the best performing architecture given the time constraint for 12000s}
Population = []\;
History = []\;
population\_size\;
sample\_size\;
pool\_size\;
\For{$1,2,\cdots,$population\_size}{
        model = random.arch()\;
        model.acc, model.time\_cost = eval(model)\;
        Population.append(model)\;
        History.append(model)\;
}
\While{not exceeding time budget}{
    Sample = []\;
    \For{$1,2,\cdots,$sample\_size}{
        Sample.append(random.choice(Population))
    }
    parent = max\_acc(Sample)\;
    GradSign\_pool = []\;
    \For(\tcc*[f]{\algname assisted part}){$1,2,\cdots,$pool\_size}{
        model = mutate\_arch(parent)\;
        model.score = GradSign(model)\;
        GradSign\_pool.append(model)\;
    }
    child = max\_score(GradSign\_pool)\;
    Population.append(child)\;
    History.append(child)\;
    Population.popleft()\;
}
\textbf{return} max\_acc(History)
 \caption{G-REA}
 \label{alg:G-REA}
\end{algorithm}

\begin{algorithm}[]
\SetAlgoLined
\KwResult{Find the best performing architecture given the time constraint for 12000s}
History = []\;
pool\_size\;
\While{not exceeding time budget}{
    GradSign\_pool = []\;
    \For(\tcc*[f]{\algname assisted part}){$1,2,\cdots,$pool\_size}{
        model = random\_arch()\;
        model.score = GradSign(model)\;
        GradSign\_pool.append(model)\;
    }
    arch = max\_score(GradSign\_pool)
    arch.acc = eval(arch)
    History.append(arch)\;
}
 \textbf{return} max\_acc(History)\;
 \caption{G-RS}
 \label{alg:G-RS}
\end{algorithm}

\begin{algorithm}[]
\SetAlgoLined
\KwResult{Find the best performing architecture given the time constraint for 12000s}
History = []\;
pool\_size\;
policy $\pi_{\theta_0}$\;
Reward = []\;
baseline\;
\While{not exceeding time budget}{
    arch = generate\_arch($\pi_{\theta_i}$)\;
    GradSign\_pool = []\;
    \For(\tcc*[f]{\algname assisted part}){$1,2,\cdots,$pool\_size}{
        child = mutate\_arch(arch)\;
        child.score = GradSign(child)\;
        GradSign\_pool.append(child)\;
    }
    arch = max\_score(GradSign\_pool)\;
    arch.acc = eval(arch)\;
    r= arch.acc\;
    History.append(arch)\;
    Reward.append(r)\;
    baseline.update(r)\;
    $\theta_{i+1} = \theta_i + \nabla_{\theta}\mathbb{E}_{\pi_{\theta_i}}[r-\text{baseline}]$
}
 \textbf{return} max\_acc(History)\;
 \textbf{return} 
 \caption{G-REINFORCE}
 \label{alg:G-REINFORCE}
\end{algorithm}

\begin{algorithm}[]
\SetAlgoLined
\KwResult{Find the best performing architecture given the time constraint for 12000s}
\textbf{Input:} budgets $b_{\min}$ and $b_{\max}$, $\eta$\;
$s_{\max} = \lfloor \log_{\eta}\frac{b_{\max}}{b_{\min}} \rfloor$\;
score\_list = []\;
pool\_size\;
\For{$s \in \{s_{\max}, s_{\max-1}, \cdots, 0\}$}{
    config\_space = []\;
    set $n=\lceil \frac{s_{\max+1}}{s+1}\cdot \eta^s \rceil$\;
    \While{sizeof(config\_space) < n}{
        GradSign\_pool = []\;
        \For(\tcc*[f]{\algname assisted part}){$1,2,\cdots,$pool\_size}{
            model = random\_arch()\;
            \eIf{model in score\_list}{
                model.score = score\_list[model]\;
            }
            {
                model.score = GradSign(model)\;
                score\_list[model] = model.score\;
            }
            GradSign\_pool.append(model)\;
        }
        arch = max\_score(GradSign\_pool)\;
        config\_space.append(arch)\;
    }
    run SH on them with initial budget as $\eta^s\cdot b_{\max}$\;
}
 \textbf{return} best evaluated architecture\;
 \caption{G-HB}
 \label{alg:G-HB}
\end{algorithm}



\end{document}